\documentclass[10pt,journal,compsoc]{IEEEtran}
\usepackage{graphicx}
\usepackage{comment}
\usepackage{amsmath,amssymb}        %
\usepackage{color}

\usepackage{xspace}
\usepackage{bbm}
\usepackage{makecell}
\usepackage{subfigure}

\usepackage{graphicx}
\usepackage{amsmath}
\usepackage{amssymb}
\usepackage{dblfloatfix}
\usepackage{transparent}
\usepackage{booktabs}
\usepackage{tabulary}
\usepackage[dvipsnames]{xcolor}
\newcommand{\method}{{SOLOv2}\xspace}
\newcommand{\OurMethod}{{SOLOv2}\xspace}

\newcommand{\myparagraph}[1]{{ \noindent \bf #1}}

\newcommand{\ie}{\textit{i}.\textit{e}.}
\newcommand{\eg}{\textit{e}.\textit{g}.}

\usepackage{algorithm}
\usepackage{listings}
\usepackage{algorithmic}
\usepackage{balance}
\usepackage{nicefrac}               %

\def\X{{$\times$}}

\usepackage[british,english,american]{babel}

\usepackage{etoolbox}
\makeatletter
\AfterEndEnvironment{algorithm}{\let\@algcomment\relax}
\AtEndEnvironment{algorithm}{\kern2pt\hrule\relax\vskip3pt\@algcomment}
\let\@algcomment\relax
\newcommand\algcomment[1]{\def\@algcomment{\footnotesize#1}}
\renewcommand\fs@ruled{\def\@fs@cfont{\bfseries}\let\@fs@capt\floatc@ruled
  \def\@fs@pre{\hrule height.8pt depth0pt \kern2pt}%
  \def\@fs@post{}%
  \def\@fs@mid{\kern2pt\hrule\kern2pt}%
  \let\@fs@iftopcapt\iftrue}
\makeatother

\definecolor{codegreen}{rgb}{0,0.5,0}
\definecolor{codeblue}{rgb}{0.25,0.5,0.5}
\definecolor{codegray}{rgb}{0.6,0.6,0.6}

\usepackage{pifont}%
\newcommand{\cmark}{\ding{51}}%
\newcommand{\xmark}{\ding{55}}%
\usepackage{hyperref}       %
\usepackage{url}            %

\usepackage[nocompress]{cite}

\usepackage{enumitem}
\setitemize{noitemsep,topsep=3pt,parsep=0pt,partopsep=0pt}

\usepackage[margin=7pt,font=footnotesize,labelfont={sc,bf},labelsep=endash,tableposition=top]{caption}

\usepackage{mathtools}
\usepackage{bbm,nicefrac}

\hypersetup{
    colorlinks=true,
    breaklinks=true,
    urlcolor= blue,
    linkcolor= red,
    bookmarksopen=false,
    filecolor=black,
    citecolor=blue,
    linkbordercolor=blue
}

\begin{document}

\title{
SOLO: A Simple Framework for \\ Instance Segmentation
}

\author{Xinlong Wang, ~~~  Rufeng Zhang, ~~~ Chunhua Shen,
~~~ Tao Kong, ~~~ Lei Li%
\IEEEcompsocitemizethanks{\IEEEcompsocthanksitem
XW, and CS are with The University of Adelaide. CS is with Monash University.
RZ is with Tongji University. TK and LL are with ByteDance AI Lab.
\IEEEcompsocthanksitem CS and TK are the corresponding authors.
}%
\thanks{\today{}.}
}

\IEEEtitleabstractindextext{%
\begin{abstract}
Compared to many other dense prediction tasks, \eg, semantic segmentation, it is the arbitrary number of instances that has made instance segmentation much more challenging.
In order to predict a mask for each instance, mainstream approaches either follow
the ``detect-then-segment'' strategy (\eg, Mask R-CNN), or predict embedding vectors first then cluster pixels into
individual instances.
In this paper, we view the task of instance segmentation from
a completely new perspective by introducing the notion of ``instance
categories'', which assigns categories to each pixel within an instance
according to the instance’s location.
With this notion, we propose \textbf{s}egmenting \textbf{o}bjects by \textbf{lo}cations (SOLO),  a simple, direct, and fast framework for instance segmentation with strong performance.

We derive %
a few SOLO variants
(\eg, Vanilla SOLO, Decoupled SOLO,  Dynamic SOLO)
following the basic principle.
Our method directly maps a raw input image to the desired object categories and instance masks, eliminating the
need for the grouping post-processing  or the bounding box detection.
Our approach achieves state-of-the-art results for instance segmentation in terms of both speed and accuracy, while being considerably simpler than the existing methods.
Besides instance segmentation, our method yields state-of-the-art results in object
detection (from our mask byproduct) and panoptic segmentation.
We further demonstrate the flexibility and high-quality segmentation of SOLO by extending it to perform one-stage instance-level image matting.
Code
is
available at: \url{https://git.io/AdelaiDet}

\end{abstract}

}

\maketitle

\section{Introduction}

\begin{figure*}[t]
\centering
    \includegraphics[width=0.8596\linewidth]{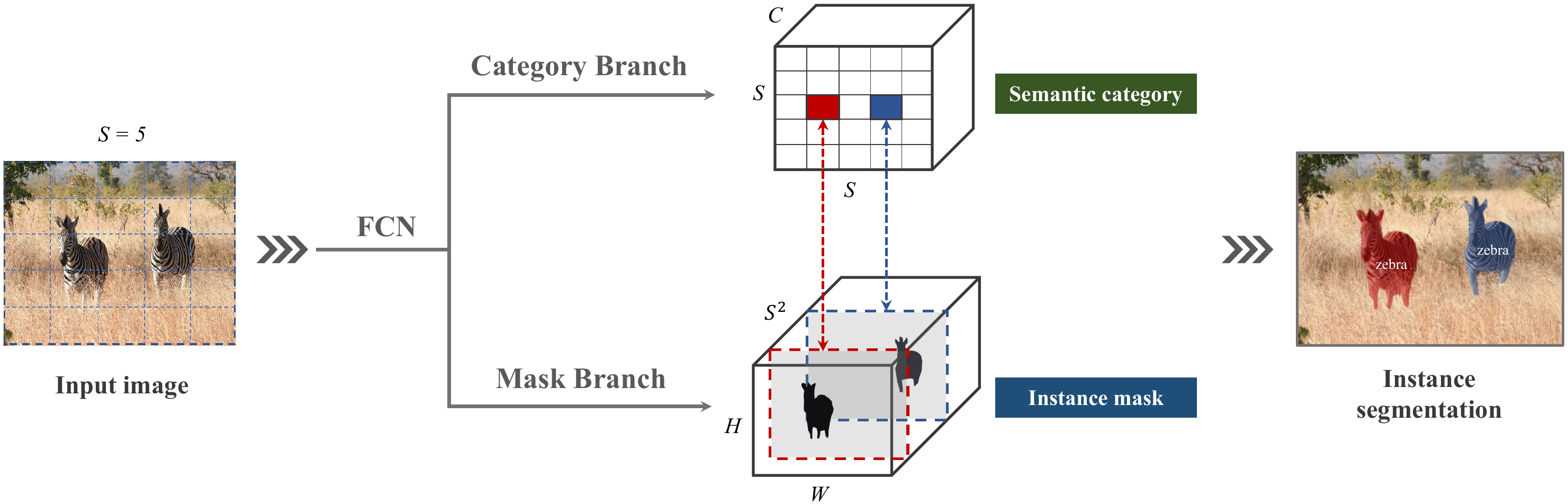}
   \caption{\textbf{SOLO framework.}
    We reformulate the instance segmentation as two sub-tasks:
    category prediction and instance mask generation problems.
    An input image is divided into a uniform grids, \ie, $S$$\times$$S$.
    Here, we illustrate the grid with $S = 5$.
    If the center of an object falls into a grid cell, that grid cell
    is responsible for predicting the semantic category (top)
    and masks of instances (bottom).
    We do not show the
    feature pyramid network (FPN) here for simpler
    illustration.
   }
\label{fig:solo_framework}
\end{figure*}

The goal of general object detection is to localize individual objects and recognize their categories.
Bounding box stands out for its simplicity for representing the object location.
Localizing objects using bounding boxes have been extensively explored,
including the problem formulation~\cite{fasterrcnn, focalloss, extremenet, reppoints, fcos}, network architecture~\cite{fpn, refinedet, tan2020efficientdet}, post-processing~\cite{bodla2017soft, he2019bounding, CaiZWLFAC19} and all those focusing on optimizing and processing the bounding boxes~\cite{yu2016unitbox, rezatofighi2019generalized, hu2018relation}.
The tailored solutions largely boost the performance and efficiency, thus enabling wide
downstream applications recently.
However, bounding boxes are coarse and unnatural.
Human vision can effortlessly localize objects by their irregular boundaries.
Instance segmentation, \ie, localizing objects using masks, pushes object localization to the limit at pixel level and opens up opportunities to more instance-level perception and applications.

Instance segmentation is challenging because it requires the correct separation of
all objects in an image while also semantically segmenting each instance at the
pixel level.
To tackle this problem, recent methods can be categorized into two groups,
\ie, top-down~\cite{fcis, maskrcnn, chen2019hybrid, Chen_2019_ICCV} and
bottom-up~\cite{associativeembedding, de2017semantic, SGN17, Gao_2019_ICCV} paradigms.
The former approach, namely `detect-then-segment', first detects bounding boxes and then
segments the instance mask in each bounding box. The latter approach learns
an affinity relation, assigning an embedding vector to each pixel, by pushing
away pixels belonging to different instances and pulling close pixels in the same
instance. A grouping post-processing is then needed to separate instances.
Both these two paradigms are step-wise and indirect,
which either heavily rely on accurate bounding box detection or depend on per-pixel embedding learning and
the grouping processing.

In contrast, we aim to directly segment instance masks,
under the supervision of full instance mask annotations instead of masks in boxes or additional pixel pairwise
relations. We start by rethinking a question:
\textit{What are the fundamental differences between object instances in an image?}
Take the challenging MS COCO dataset~\cite{coco} for example.
There are in total $36,780$ objects in the validation subset, $98.3\%$ of object pairs have center distance greater than $30$ pixels.
As for the rest $1.7\%$ of object pairs, $40.5\%$ of them have size ratio greater than 1.5$\times$.
To conclude, in most cases, two instances in an image either have different center locations or have different object sizes.
This observation makes one wonder whether we could directly distinguish instances by the center locations and object sizes?

In the closely related field, semantic segmentation,  now
the dominated paradigm leverages a fully convolutional network (FCN) to output dense predictions with $N$ channels. Each output channel is responsible for one of the semantic categories (including background).
Semantic segmentation aims to distinguish different semantic categories.
Analogously,
in this work, we propose to distinguish object instances
in the image  by introducing the notion of ``\textit{instance categories}'', \ie, the quantized center locations and object sizes, which enables to \textbf{s}egment \textbf{o}bjects by \textbf{lo}cations, thus the name of our framework, SOLO.

For center locations, an image can be divided into a grid of $S \times S$ cells, thus leading to $S^2$ center location classes. According to the coordinates of the object center, an object instance is assigned to one of the grid cells, as its center location category.
Note that grids are used conceptually to assign a location category for each pixel.
Each output channel is responsible for one of the center location categories, and the corresponding channel map should predict the instance mask of the object belonging to that location.
Thus,  structural geometric information is naturally preserved in the spatial matrix with dimensions of height by width.
Unlike DeepMask~\cite{deepmask} and TensorMask~\cite{Chen_2019_ICCV}, which run in a dense sliding-window manner and segment an object in a fixed local patch, our method
naturally outputs accurate masks for all scales of instances without the limitation of (anchor) box locations and scales.
In essence, an instance location category approximates
the location of the object center of an instance.
Thus, by classification of each pixel into its instance location category, it is equivalent to predict the object center of each pixel in the latent space.
The importance here of converting
the location prediction task into classification is that, with classification it is much
more straightforward and easier to model
a varying number of instances using a fixed number of channels, at the same time not relying on post-processing like grouping or learning embeddings.
For object sizes, we employ the feature pyramid network (FPN)~\cite{fpn} to distinguish instances with different object sizes, so as to
assign objects of different sizes to different levels of feature maps.
Note that FPN was designed for the purposes of detecting objects of different sizes in an image.
Thus, all the object instances are separated regularly, enabling to classify objects by ``instance categories''.

With the proposed SOLO framework, we are able to optimize the network in
an end-to-end fashion for the instance segmentation task and perform pixel-level instance segmentation out of the restrictions of local box detection and pixel grouping.
To fully exploit the capabilities of this simple framework, we propose three different SOLO variants following the basic principle, namely vanilla SOLO, Decoupled SOLO and Dynamic SOLO (SOLOv2).

Besides the problem formulation, the supporting facilities of instance segmentation, \eg, post-processing,
is largely unexplored compared to bounding box detection.
For developing a fast and pure instance segmentation framework, we propose an efficient and effective matrix NMS algorithm.
As a post-processing step for suppressing the duplicate predictions, non-maximum suppression (NMS) serves as an integral part in state-of-the-art object detection systems.
Take the widely adopted multi-class NMS for example.
For each class, the predictions are sorted in descending order by confidence.
Then for each prediction, it removes all other highly overlapped predictions.
Such sequential and recursive operations result in non-negligible latency.
For mask NMS, this drawback is further magnified.
Compared to bounding box, it consumes more time to compute the IoU of each mask pair, thus leading to huge overhead.
We address this problem by introducing Matrix NMS, which performs NMS with parallel matrix operations in one shot.
Our Matrix NMS outperforms the existing NMS and its varieties in both accuracy and speed.
As a result, \textit{Matrix NMS processes 500 masks in less than 1 ms in simple python implementation}, and outperforms the recently proposed Fast NMS~\cite{yolact} by 0.4\%  AP.

To summarize, our simple framework outperforms the state-of-the-art instance segmentation methods in both speed and accuracy.
Our model with ResNet-50 backbone achieves 38.8\% mask AP at 18 FPS on the challenging MS COCO dataset,
evaluated on a single V100 GPU card. A light-weight version executes at 31.3 FPS and
yields 37.1\% mask AP.
Interestingly, although the concept of bounding box is thoroughly eliminated
in our method, our bounding box byproduct, \ie, by directly converting the predicted mask to its bounding box, yields 44.9\% AP for object detection, which even surpasses many state-of-the-art, highly-engineered object detection methods.
By adding the semantic segmentation branch, we easily extend our method to solve panoptic segmentation and achieve state-of-the-art results.
We also extend our framework to perform image matting at instance-level.
Thanks to the ability of generating high-quality object masks, our method is able to solve instance-level image matting in one shot with minimal modifications.

We believe that, with our simple, fast and sufficiently strong solution, instance segmentation can be a popular alternative to the widely used object bounding box detection, and SOLO may play an important role and predict its wide applications.

\section{Related Work}
We review some works that are closest to ours.

\myparagraph{Top-down instance segmentation.}
The methods that segment object instances in a priori bounding box fall into the typical top-down paradigm.
FCIS~\cite{fcis} assembles the position-sensitive score maps within the region-of-interests (ROIs) generated by a region proposal network (RPN) to predict instance masks.
Mask R-CNN~\cite{maskrcnn} extends the Faster R-CNN detector~\cite{fasterrcnn} by adding a branch for segmenting the object instances within the detected bounding boxes.
Based on Mask R-CNN, PANet~\cite{panet} further enhances the feature representation to improve the accuracy,
Mask Scoring R-CNN~\cite{maskscoringrcnn} adds a mask-IoU branch to predict the quality of the predicted mask for improving the performance.
HTC~\cite{chen2019hybrid} interweaves box and mask branches for a joint multi-stage processing.
TensorMask~\cite{Chen_2019_ICCV} adopts the dense sliding window paradigm to segment the instance in the local window for each pixel with a predefined number of windows and scales.
YOLACT~\cite{yolact} learns a group of coefficients which are normalized to ($-1$, 1) for each anchor box. During the inference, it first performs a bounding box detection and then uses the predicted boxes to crop the assembled masks.
Our Dynamic SOLO directly decouples the original mask prediction
into kernel learning and feature learning.
No anchor box is needed. No normalization is needed. No bounding box detection is needed.
Both the training and inference are much simpler.
In contrast to the top-down methods above, our SOLO framework is totally box-free thus not being restricted by (anchor) box locations and scales, and naturally benefits from the inherent advantages of FCNs.

\myparagraph{Bottom-up instance segmentation.}
This category  of the approaches generate instance masks by grouping the pixels into an arbitrary number of object instances presented in an image.
In~\cite{associativeembedding}, pixels are grouped into instances using the learned associative embedding.
A discriminative loss function~\cite{de2017semantic} learns pixel-level instance embedding efficiently, by pushing away pixels belonging to different instances and pulling close pixels in the same instance.
SGN~\cite{SGN17} decomposes the instance segmentation problem into a sequence of sub-grouping problems.
SSAP~\cite{Gao_2019_ICCV} learns a pixel-pair affinity pyramid, the probability that two pixels belong to the same instance, and sequentially generates instances by a cascaded graph partition.
Typically,
 bottom-up methods lag behind in accuracy compared to top-down methods, especially on the dataset with diverse scenes.
Instead of exploiting pixel pairwise relations
and pixel grouping,
our method directly learns %
with the instance mask annotations solely during training,
and predicts instance masks end-to-end without
grouping post-processing.

\myparagraph{Direct instance segmentation.}
To our knowledge, no prior methods directly train with
mask
annotations solely,  and predict instance masks and semantic categories in one shot without the need of grouping post-processing.
Several recently proposed methods may be viewed as the `semi-direct' paradigm.
AdaptIS~\cite{adaptis} first predicts point proposals, and then sequentially generates the mask for the object located at the detected point proposal.
PolarMask~\cite{polarmask} proposes to use the polar representation
to encode masks
and transforms per-pixel mask prediction to distance regression.
They both do not need bounding boxes for training but are either being step-wise or founded on compromise, \eg, coarse parametric representation of masks.
Our SOLO framework takes an image as input,
directly outputs instance masks and corresponding class probabilities,
in a fully convolutional, box-free and grouping-free paradigm.

\myparagraph{Dynamic convolutions.}
In traditional convolution layers, the learned convolution kernels stay fixed and are independent on the input, \ie, the weights are the same for arbitrary image and any location of the image.
Some previous works~\cite{jaderberg2015spatial,jia2016dynamic,dai2017deformable} explore the idea of bringing more flexibility into the traditional convolutions.
Dynamic filter~\cite{jia2016dynamic} is proposed to actively predict the parameters of the convolution filters. It applies dynamically generated filters to an image in a sample-specific way.
Deformable Convolutional Networks~\cite{dai2017deformable} dynamically learn the sampling locations by  predicting the offsets for each image location.
Pixel-adaptive convolution~\cite{SuJSGLK19} multiplies the weights of the filters and a spatially varying kernel to make the standard convolution content-adaptive.
In the enhanced version SOLOv2, we bring the dynamic scheme into instance segmentation and enable learning instance segmenters by locations.
Note that the
work in \cite{CondInst} also applies dynamic convolutions for instance segmentation by extending the framework of BlendMask \cite{chen2020blendmask}.
The dynamic scheme part is somewhat similar, but the methodology is different. CondInst \cite{CondInst}   relies on the relative position to distinguish instances as in AdaptIS, while SOLOv2 uses absolute positions as in SOLO. It means that the former needs to encode the position information $N$ times for $N$ instances, while \method performs it all at once using the global coordinates, regardless how many instances there are.

\myparagraph{Non-maximum suppression.}
NMS is widely adopted in many computer vision tasks and becomes an essential component of object detection
and instance segmentation
systems.
Some recent works~\cite{bodla2017soft, liu2019adaptive, he2019bounding, CaiZWLFAC19, yolact} are proposed to improve the traditional NMS.
They can be divided into two groups, either for improving the accuracy or speeding up.
Instead of applying the hard removal to duplicate predictions according to a threshold,  Soft-NMS~\cite{bodla2017soft} decreases the confidence scores of neighbors according to their overlap with higher scored predictions.
In~\cite{he2019bounding}, the authors use KL-Divergence and reflected it in the refinement of coordinates in the NMS process.
To accelerate the inference, Fast NMS~\cite{yolact} enables deciding the predictions to be kept or discarded in parallel.
Note that
it speeds up at the cost of performance
deterioration.
Different from the previous methods, our Matrix NMS addresses the issues of hard removal and sequential operations at the same time.
As a result, \textit{the proposed Matrix NMS is able to process 500 masks in less than 1 ms} in simple python implementation, which is
negligible compared with the time of network
evaluation,
and yields 0.4\%  AP better than Fast NMS.

\section{Our Approach}
In this section, we first introduce how our SOLO framework reformulates instance segmentation as a per-pixel classification problem. Next, we provide three different variants following the basic principle. At last, we present the learning and inference strategies, as well as the Matrix NMS.

\subsection{Problem Formulation}
The central idea of SOLO framework is to reformulate the instance segmentation as two simultaneous category-aware prediction problems.
Concretely, our system conceptually divides the input image into a uniform
grid,
\ie, $S$$\times$$S$. If the center of an object falls into a grid cell, that grid cell
is responsible for 1) predicting the semantic category as well as 2) segmenting that object instance.

\subsubsection{Semantic Category}
For each grid, SOLO predicts the $C$-dimensional output to indicate the semantic class probabilities,  where $C$ is the number of classes. These probabilities are conditioned on the grid cell. If we divide the input image into $S$$\times$$S$ grids, the output space will be  $S$$\times$$S$$\times$$C$, as shown in Figure~\ref{fig:solo_framework} (top). This design is based on the assumption that each cell of the $S$$\times$$S$ grid must belong to one individual instance, thus only belonging to one semantic category. During inference, the $C$-dimensional output indicates the class probability for each object instance.

\begin{figure*}[tb]
\centering
\subfigure[Vanilla SOLO]{
\includegraphics[width=0.45\textwidth]{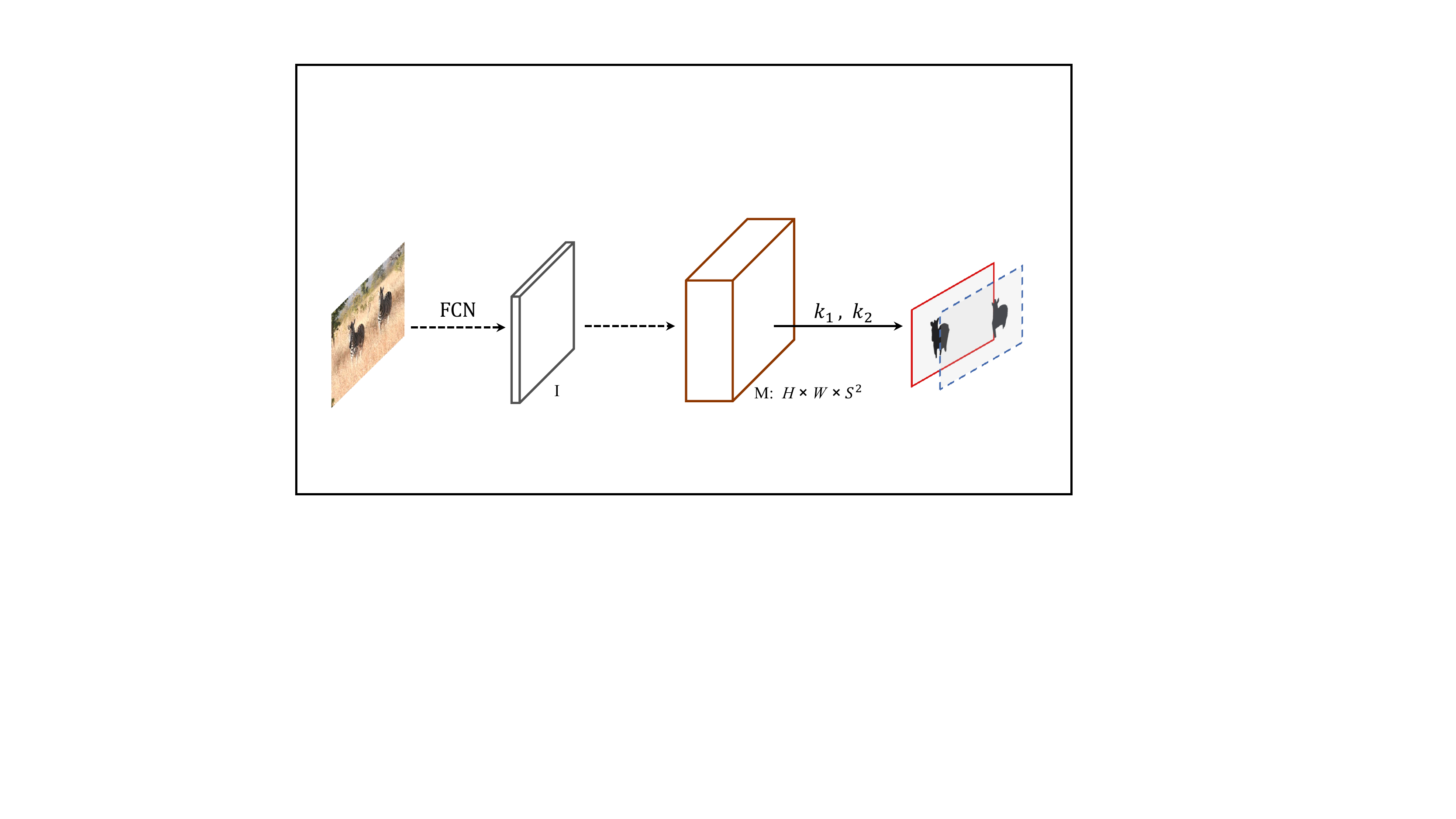}
\label{fig:vanilla_head}
}
\subfigure[Decoupled SOLO]{
\includegraphics[width=0.45\textwidth]{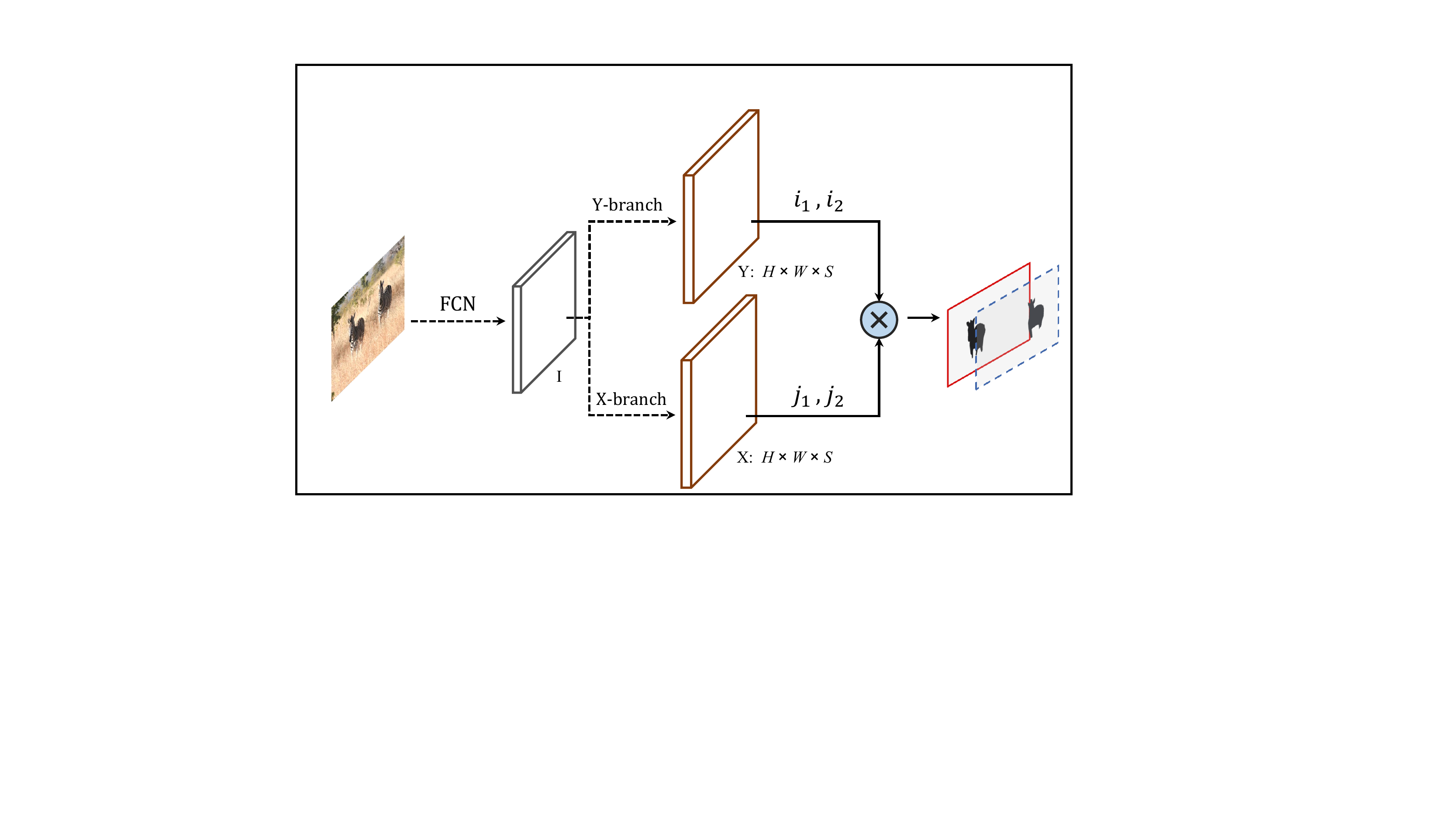}
\label{fig:decoupled_head}
}
\subfigure[Dynamic SOLO]{
\includegraphics[width=0.45\textwidth]{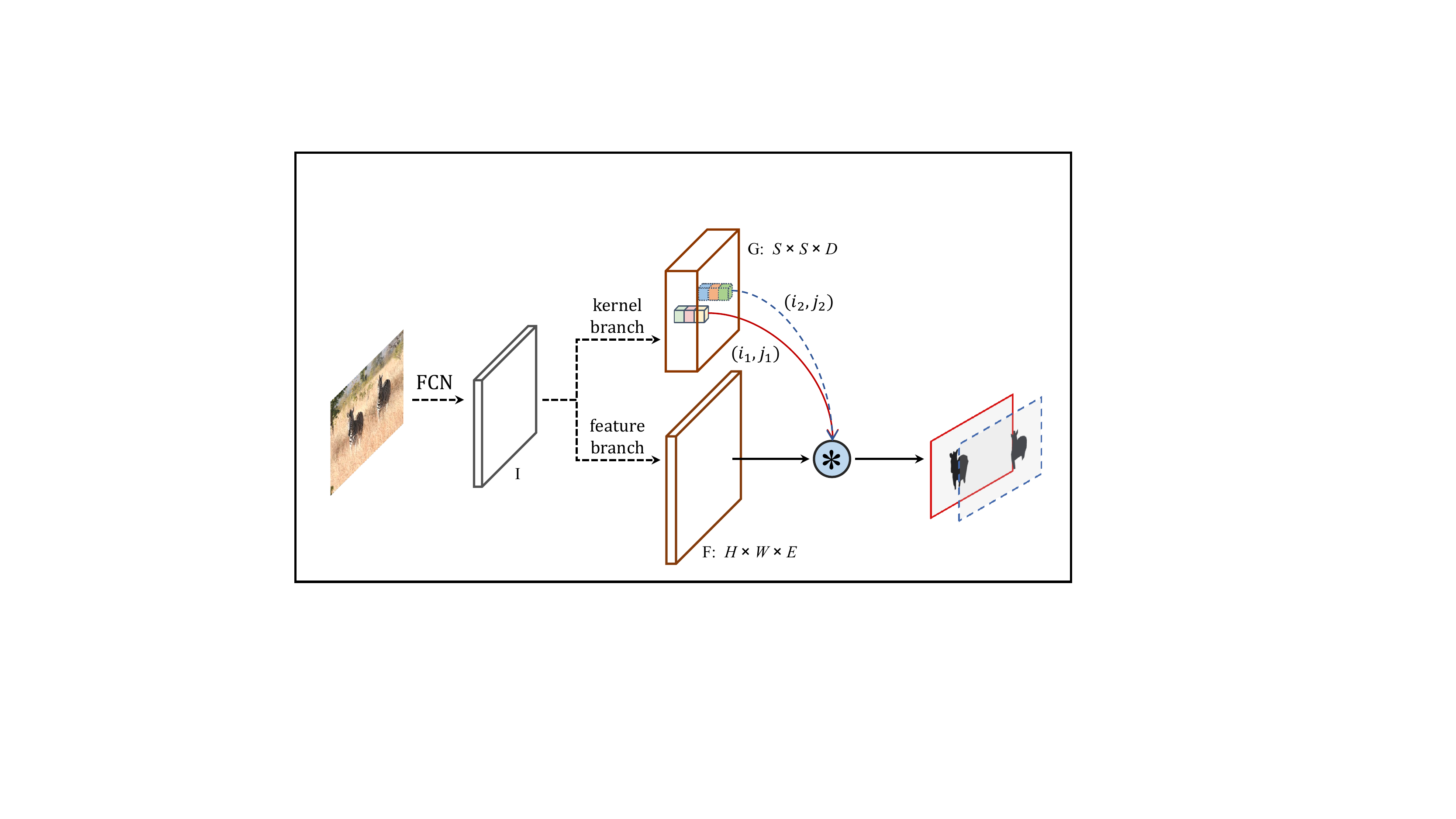}
\label{fig:dynamic_head}
}
\hspace{0.1in}
\subfigure[Decoupled Dynamic SOLO]{
\includegraphics[width=0.44\textwidth]{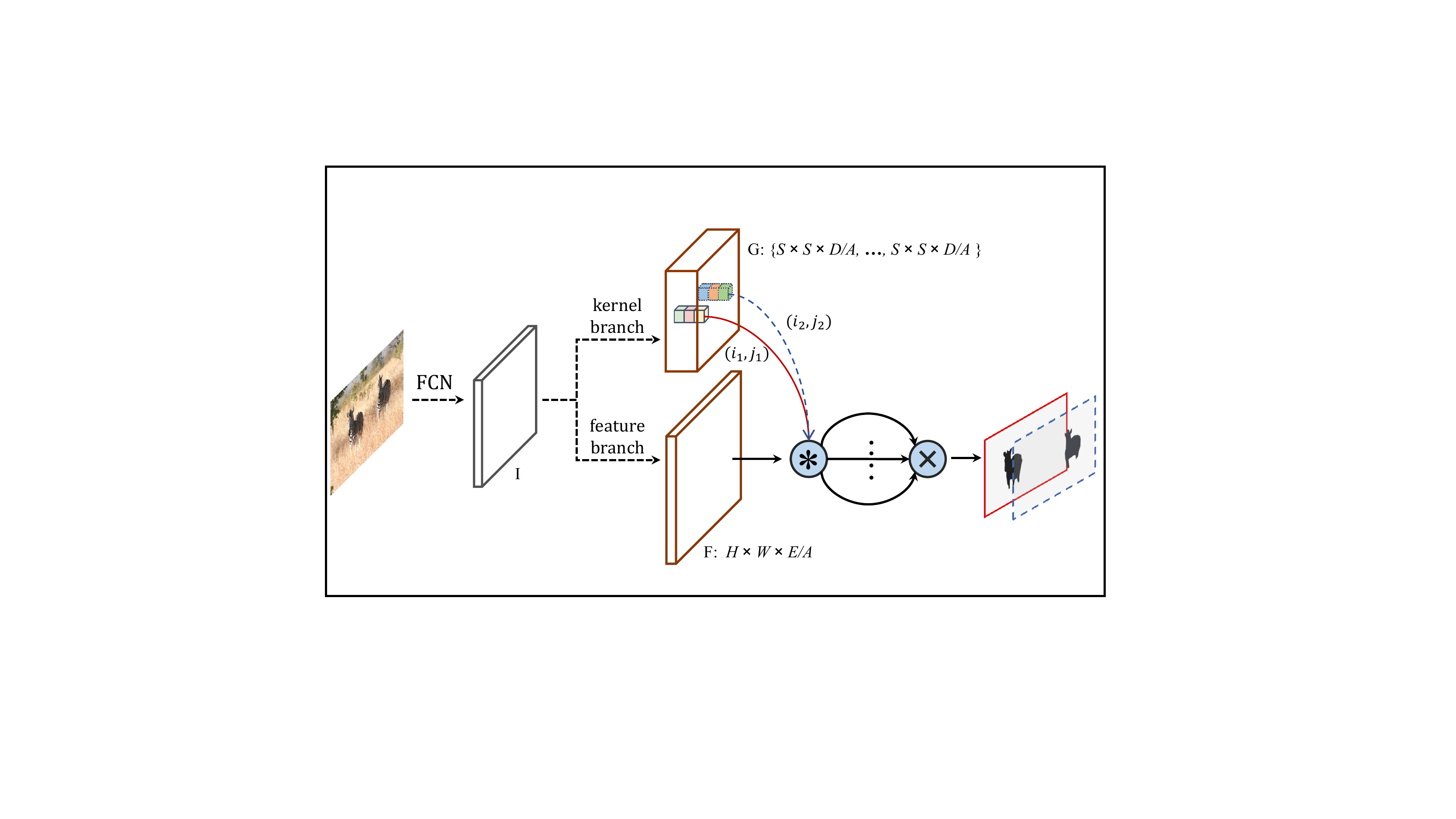}
\label{fig:dd_head}
}
\caption{\textbf{SOLO variants.} %
 $I$ is the input feature after FCN-backbone representation extraction. Black dashed arrows denote convolutions. $k=i \cdot  S + j$. ``$\otimes$'' denotes element-wise multiplication. ``$\circledast$''  denotes the dynamic convolution operation.
 }
\label{fig:variants}
\end{figure*}

\subsubsection{Instance Mask}
In parallel with the semantic category prediction, each positive grid cell will also generate the corresponding instance mask. For an input image $I$, if we divide it into $S$$\times$$S$ grids, there will be at most $S^2$ predicted masks in total. We explicitly encode these masks at the third dimension (channel) of a
3D output tensor. Specifically, the instance mask output will have $H_I$$\times$$W_I$$\times$$S^2$ dimension. The $k^{th}$ channel will be responsible to segment instance at grid ($i$, $j$), where $k=i\cdot S + j$ (with $i$ and $j$ zero-based).
 We also show %
 more efficient variants in Section \ref{sec:network_arch}.
 To this end,  a one-to-one correspondence is established between the semantic category and class-agnostic mask (Figure~\ref{fig:solo_framework}).

A direct approach to predict the instance mask is to adopt the fully convolutional networks, like FCNs in semantic segmentation~\cite{fcn}.
However, the conventional convolutional operations are \textit{spatially invariant} to some degree.
Spatial invariance is desirable for some tasks such as semantic segmentation as it enables spatially equivalent prediction.
However, here we need a model that is \textit{spatially variant},
or in more precise words, position sensitive,
since our segmentation masks are conditioned on the grid cells and must be separated by different feature channels.

Our solution is very simple:
at the beginning of the network, we directly feed normalized pixel coordinates to the
networks, inspired  by  `CoordConv' operator~\cite{coordconv}. Specifically, we create a tensor of the same spatial size as input that contains pixel coordinates,
which are  normalized to $[-1, 1]$.
This tensor is then concatenated to the input features and passed to the following layers. By simply %
giving  the convolution access to its own input coordinates, we add the spatial functionality to the conventional FCN model.
It should be noted that CoordConv is not the only choice. For example the semi-convolutional operators~\cite{semiconv}  may  be competent, but we employ CoordConv for its simplicity and being easy to implement.
If the original feature tensor is of size $H$$\times$$W$$\times$$D$, the size of the new tensor becomes $H$$\times$$W$$\times$$(D+2)$, in which the last two channels are $x$-$y$ pixel coordinates. For more information on CoordConv, we refer readers to~\cite{coordconv}.

\subsubsection{Forming Instance Segmentation}
In SOLO, the category prediction and the corresponding mask are naturally associated by their reference grid cell, \ie, $k=i\cdot S + j$. Based on this, we can directly form the final instance segmentation result for each grid. The raw instance segmentation results are generated by gathering all grid results. Finally, mask non-maximum-suppression (NMS) is used to
obtain the final instance segmentation results. No other post-processing operations are
needed.

\subsection{Network Architecture}
\label{sec:network_arch}

SOLO attaches to a convolutional backbone, \eg, ResNet \cite{resnet}.
We use FPN \cite{fpn}, which generates a pyramid of feature maps with different sizes with  a  fixed number of channels for each level.
These feature maps are used as input for the following parallel branches to generate the final predictions, \ie, semantic categories and instance masks.

To demonstrate the generality and effectiveness of our framework, we introduce the vanilla SOLO and a few efficient variants, \eg, Decoupled SOLO and Dynamic SOLO (SOLOv2).
We show the overall comparison in Figure~\ref{fig:variants}.
We note that our instance segmentation heads have a straightforward structure. More complex designs have the potential to improve performance, but are not the focus of this work.

\subsubsection{Vanilla SOLO}
As shown in Figure~\ref{fig:vanilla_head}, vanilla SOLO has the simplest architecture among all three variants.
It directly maps the input image to the output tensor $M$.
The $k_{1}^{th}$ and $k_{2}^{th}$ instance mask could be obtained from the $k_{1}^{th}$ and $k_{2}^{th}$ channel of $M$.

\begin{figure}[t!]
\begin{center}
    \includegraphics[width=0.8\linewidth]{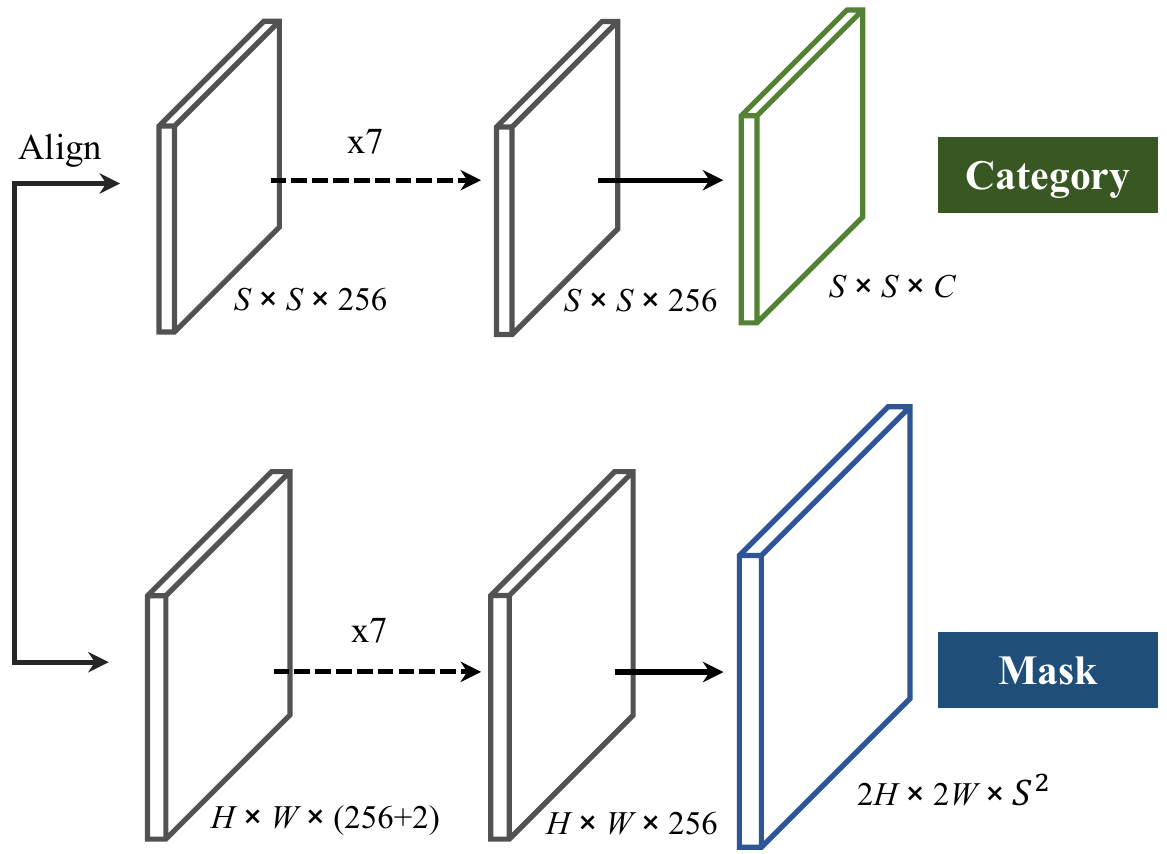}
\end{center}
   \caption{\textbf{Vanilla Head architecture}. At each FPN feature level, we attach two sibling sub-networks, one for instance category prediction (top) and one for instance mask segmentation (bottom).
   In the mask branch, we concatenate the $x$, $y$ coordinates and the original features to
   encode spatial information.
   Here, numbers denote spatial resolution and channels.
   In this figure, we assume 256 channels as an example.
   Arrows denote either convolution or interpolation.
   `Align' means
   bilinear interpolation.
   During inference, the mask branch outputs
   are further upsampled to the original image size.}
\label{fig:vanilla_head_arch}
\end{figure}

We instantiate vanilla SOLO with multiple architectures. The differences include: (a) the \textit{backbone} architecture used for feature extraction, (b) the network \textit{head} for computing the instance segmentation results, and (c) training \textit{loss function} used to optimize the model.
Most of the experiments are based on the \textit{head} architecture as shown in Figure~\ref{fig:vanilla_head_arch}.

\subsubsection{Decoupled SOLO}
Given an predefined grid number, \eg, $S=20$, vanilla SOLO head outputs $S^2 = 400$ channel maps.
However, the prediction is somewhat redundant, as in most cases the objects are located sparsely in the image.
In this section, we further introduce an equivalent and significantly more efficient variant of the vanilla SOLO, termed Decoupled SOLO, shown in Figure~\ref{fig:decoupled_head}.

In Decoupled SOLO, the original output tensor $M\in \mathbb{ R}^{H\times W\times S^2}$  is replaced with two output tensors $X\in \mathbb{ R}^{H\times W\times S}$  and $Y\in \mathbb{ R}^{H\times W\times S}$, corresponding two axes respectively.
Thus, the output space is decreased from  $H$$\times$$W$$\times$$S^2$ to  $H$$\times$$W$$\times$$2S$.
For an object located at grid location $(i, j)$,
the mask prediction of that object is defined as the element-wise multiplication of two channel maps:
\begin{equation}
\begin{aligned}
\mathbf{m}_k = \mathbf{x}_j \otimes \mathbf{y}_i,
\end{aligned}
\end{equation}
where $\mathbf{x}_j$ and $\mathbf{y}_i$ are the $j^{th}$ and $i^{th}$ channel map of $X$ and $Y$ after \texttt{sigmoid} operation.
The motivation behind this is that \textit{the probability of a pixel belonging to location category $(i, j)$ is the joint probability of belonging to $i^{th}$ row and $j^{th}$ column},
as the horizontal and vertical location categories are independent.

\subsubsection{Dynamic SOLO}
In vanilla SOLO, to generate the instance mask of $S^2$ channels corresponding to $S\times S$ grids, the last layer takes one level of pyramid features $F\in \mathbb{ R}^{H\times W\times E}$ as input and at last applies a convolution layer with $S^2$ output channels.
The operation can be written as:
\begin{equation}
\label{eq:convop}
\begin{aligned}
M_{i,j} = G_{i,j} \circledast F,
\end{aligned}
\end{equation}
where $G_{i,j}\in \mathbb{ R}^{1\times 1\times E}$ is the convolution kernel, and $M_{i,j}\in \mathbb{R}^{H\times W}$ is the final mask containing only one instance whose center is at location $(i,j)$.

In other words, we need two input $F$ and $G$ to generate the final mask $M$.
From another perspective, if we separately learn $F$ and $G$, the final $M$ could be directly generated using the both components.
In this way, we can simply pick the valid ones from predicted $S^2$ kernels and perform the convolution dynamically.
The number of model parameters also decreases.
Furthermore, as the predicted kernel is generated dynamically conditioned on the input, it benefits from the flexibility and adaptive nature.
Additionally, each of the $S^2$ kernels is conditioned on the location.
It is in accordance with the core idea of segmenting objects by locations and goes a step further by predicting the segmenters by locations.

\myparagraph{Mask kernel $G$.}
Given the backbone and FPN, we predict the mask kernel $G$ at each pyramid level.
We first resize the input feature $F_I \in \mathbb{ R}^{H_I\times W_I\times C}$ into the shape of $S\times S\times C$.
Then 4\X convs and a final $3\times 3\times D$ conv are employed to generate the kernel $G$.
We add the spatial functionality to $F_I$ by giving the first convolution access to the normalized coordinates following CoordConv~\cite{coordconv}, \ie, concatenating two additional input channels which contains pixel coordinates normalized to $[-1, 1]$. Weights for the head are shared
across different feature map levels.
For each grid, the kernel branch predicts the $D$-dimensional output to indicate predicted convolution kernel weights, where $D$ is the number of parameters.
For generating the weights of a $1$$\times$$1$ convolution with $E$ input channels, $D$ equals $E$.
As for $3$$\times$$3$ convolution, $D$ equals $9E$.
These generated weights are conditioned on the locations, \ie, the grid cells.
If we divide the input image
into $S$$\times$$S$ grids, the output space will be $S$$\times$$S$$\times$$D$, There is no activation function on the output.

\myparagraph{Mask feature $F$.}
Since the mask feature and mask kernel are decoupled and separately predicted, there are two ways to construct the mask feature.
We can put it into the head, along with the kernel branch. It means that we predict the mask features for each FPN level.
Or, to predict a unified mask feature representation for all FPN levels.
We have compared the two implementations in Section~\ref{subsubsec:ablation2} by experiments.
Finally, we employ the latter one for its effectiveness and efficiency.
For learning a
unified and high-resolution mask feature representation, we apply  feature pyramid fusion inspired by the semantic segmentation in~\cite{kirillov2019panoptic}.
After repeated stages of $3\times3$ conv., group norm~\cite{wu2018group}, ReLU and $2\times$ bilinear upsampling, the FPN features P2 to P5 are merged into a single output at $\nicefrac{1}{4} $  scale.
The last layer after the element-wise summation consists of $1 \times 1$ convolution, group norm and ReLU.
It should be noted that we
feed normalized pixel coordinates to the deepest FPN level (at
$\nicefrac{1}{32} $  scale), before the convolutions and bilinear upsamplings.
The provided accurate position information is important for enabling position sensitivity and predicting instance-aware features.

\subsubsection{Decoupled Dynamic SOLO}
We further develop a more unified algorithm by combining decoupled SOLO~\cite{solo} and dynamic SOLO~\cite{wang2020solov2}.
Specifically, based on dynamic SOLO, we further decouple the mask representation inspired by decoupled SOLO.
We divide the predicted mask kernel weights $G  \in \mathbb{ R}^{S\times S\times D}$ into $A$ groups, \ie, $G_1, ..., G_A$. Each group is in the shape of ${S\times S\times \nicefrac{D}{A}}$.
The mask feature $F$ is reduced from $E$ channels to $\nicefrac{E}{A}$ channels.
For an object, its mask is predicted as:
\begin{equation}
\begin{aligned}
M &= M_1 \otimes ... \otimes M_A \\
&= \texttt{sigmoid}(G_1 \circledast F) \otimes ...  \otimes \texttt{sigmoid}(G_A \circledast F).
\end{aligned}
\end{equation}
Decoupled mask representation enables to maintain high
performance with a compact mask feature.
As such, we adopt mask feature $F$ with reduced output space, \eg, from
$H\times W\times 256$ to $H\times W\times 64$ for $A = 4$.
The frameworks of the proposed SOLO variants are illustrated and compared in Figure~\ref{fig:variants}.

\subsection{Learning and Inference}

\subsubsection{Label Assignment}
For the category prediction branch, the network needs to give the object category probability for each of $S$$\times$$S$ grid. Specifically, grid $(i, j)$ is considered as a positive sample if it falls into the \textit{center region} of any ground truth mask, Otherwise it is a negative sample. Center sampling is effective in recent works of object detection~\cite{fcos,foveabox}, and here we also utilize a similar technique for mask category classification. Given the mass center $(c_x, c_y)$, width $w$, and height $h$ of the ground truth mask, the center region is controlled by constant scale factors $\epsilon$: $(c_x, c_y, \epsilon w, \epsilon h)$. We set $\epsilon=0.2$ and there are on average 3 positive samples for each ground truth mask.

Besides the label for instance category, we also have a binary segmentation mask for each positive sample. Since there are $S^2$ grids, we also have $S^2$ output masks for each image. For each positive samples, the corresponding target binary mask will be annotated. One may be concerned that the order of masks will impact the mask prediction branch, however, we show that the most simple row-major order works well for our method.

\subsubsection{Loss Function}
We define our training loss function as follows:
\begin{equation}
\label{eq:loss_all}
\begin{aligned}
L = L_{cate} + \lambda L_{mask},
\end{aligned}
\end{equation}
where $L_{cate}$ is the conventional Focal Loss~\cite{focalloss} for semantic category classification. $L_{mask}$ is the loss for mask prediction:
\begin{equation}
\label{eq:loss}
\begin{aligned}
L_{mask}  = \frac{1}{N_{pos}} \sum_{k}  \mathbbm{1}_{\{\mathbf{p}^{*}_{ i,j  } > 0\}} d_{mask} (\mathbf{m}_{k}, \mathbf{m}^{*}_{k}),
\end{aligned}
\end{equation}
Here indices $ i =  \lfloor k/S\rfloor, j = k \, {\rm mod} \, S $,
if we index the grid cells  (instance category labels) from left to right and top to down.
$N_{pos}$ denotes the number of positive samples, $\mathbf{p}^{*}$ and $\mathbf{m}^{*}$ represent category and mask target
respectively.
$\mathbbm{1}$ is the indicator function, being 1 if $\mathbf{p}^{*}_{i, j} > 0$ and 0 otherwise.

We have compared different implementations of $d_{mask}(\cdot,\cdot)$: Binary Cross Entropy (BCE), Focal Loss~\cite{focalloss} and Dice Loss~\cite{vnet}. Finally, we employ Dice Loss for its effectiveness and stability in training.
$\lambda$ in Equation~\eqref{eq:loss_all} is set to 3.
The Dice Loss is defined as
\begin{equation}
\begin{aligned}
L_{Dice} = 1 - D(\mathbf{p}, \mathbf{q}),
\end{aligned}
\end{equation}
where $D$ is the dice coefficient, which is defined as
\begin{equation}
\begin{aligned}
D(\mathbf{p}, \mathbf{q}) = \frac{2\sum_{x,y}(\mathbf{p}_{x,y} \cdot  \mathbf{q}_{x,y})}{\sum_{x,y}\mathbf{p}^2_{x,y} + \sum_{x,y}\mathbf{q}^2_{x,y}}.
\end{aligned}
\end{equation}
Here $\mathbf{p}_{x,y}$ and $\mathbf{q}_{x,y}$ refer to the value of pixel located at $(x, y)$ in
predicted soft mask $\mathbf{p}$ and ground truth mask $\mathbf{q}$.

\subsubsection{Inference Pipeline}

During the inference,
we forward input image through the backbone network and FPN, and obtain the category score
$\mathbf{p}_{i,j}$ at grid $(i, j)$.
We first use a confidence threshold of $0.1$ to filter out predictions with low confidence.
For vanilla and decoupled SOLO, we select the top $500$ scoring masks and feed them into the NMS operation.
For dynamic SOLO, the corresponding predicted mask kernels are used to perform convolution on the mask feature to generate the soft masks.
After the \texttt{sigmoid} operation, we use a threshold of $0.5$ to convert predicted soft masks to binary masks.
The last step is the Matrix NMS.

\myparagraph{Maskness.} We calculate maskness for each predicted mask, which represents the quality and confidence of mask prediction ${\tt  maskness} = \frac{1}{N_{f}} \sum_{i}^{N_{f}} {\mathbf{p}_{i}}$.
Here $N_f$ is the number of foreground pixels of the predicted soft mask $\mathbf{p}$, \ie, the pixels that have values greater than threshold $0.5$.
The classification score for each prediction is multiplied by the mask-ness as the final confidence score.

\def\iou{{{\tt iou}}}
\subsubsection{Matrix NMS}

\myparagraph{Motivation.}
Our Matrix NMS is motivated by Soft-NMS~\cite{bodla2017soft}. Soft-NMS decays the other detection scores as a monotonic decreasing function $f(\iou)$ of their overlaps. By decaying the scores according to IoUs recursively, higher IoU detections will be eliminated with a minimum score threshold. However, such process is sequential like traditional Greedy NMS and could not be implemented in parallel.

Matrix NMS views this process from another perspective by considering how a predicted mask $m_j$ being suppressed. For $m_j$, its decay factor is affected by: (a) The penalty of each prediction $m_{i}$ on $m_j$ ($s_i>s_j$), where $s_i$ and $s_j$ are the confidence scores; and (b) the probability of $m_i$ being suppressed. For (a), the penalty of each prediction $m_{i}$ on $m_j$ could be easily computed by $f(\iou_{i,j})$. For (b), the probability of $m_i$ being suppressed is not so elegant to be computed. However, the probability usually has positive correlation with the IoUs. So here we directly approximate the probability by the most overlapped prediction on $m_i$ as
\begin{equation}
\label{eq:prob}
    f(\iou_{\cdot, i}) = \min_{\forall s_k>s_i}f(\iou_{k, i}).
\end{equation}
To this end, the final decay factor becomes
\begin{equation}
\label{eq:matrixnms}
    decay_j = \min_{\forall s_i>s_j} \frac{f(\iou_{i, j})}{f(\iou_{\cdot, i})},
\end{equation}
and the updated score is computed  by $s_j = s_j \cdot decay_j$.
We consider two most simple decremented functions, denoted as \texttt{linear} %
$
    f(\iou_{i,j}) = 1 - \iou_{i,j},
$
and \texttt{Gaussian}
$
    f(\iou_{i,j})= \exp \Big( {-\frac{\iou_{i,j}^2}{\sigma}}
                       \Big).
$

\myparagraph{Implementation.} All the operations in Matrix NMS could be implemented in one shot without recurrence. We first compute a $N\times N$ pairwise IoU matrix for the top $N$ predictions sorted descending by score. For binary masks, the IoU matrix could be efficiently implemented by matrix operations. Then we get the most overlapping IoUs by column-wise max on the IoU matrix. Next, the decay factors of all higher scoring predictions are computed, and the decay factor for each prediction is selected as the most effect one by column-wise min (Equation~\eqref{eq:matrixnms}). Finally, the scores are updated by the decay factors. For usage, we %
only
need thresholding and selecting top-$k$ scoring masks as the final predictions.

The pseudo-code of Matrix NMS is provided in Figure~\ref{fig:matrix_nms}. In our code base, Matrix NMS is 9$\times$  faster than traditional NMS and being more accurate (Table~\ref{tab:nms}\textcolor{red}{(c)}).
We show that Matrix NMS serves as a superior alternative of traditional NMS in both  accuracy and speed, and can be easily integrated into the state-of-the-art detection/segmentation systems.

\lstset{
  backgroundcolor=\color{white},
  basicstyle=\fontsize{7.5pt}{8.5pt}\fontfamily{lmtt}\selectfont,
  columns=fullflexible,
  breaklines=true,
  captionpos=b,
  commentstyle=\fontsize{8pt}{9pt}\color{codegray},
  keywordstyle=\fontsize{8pt}{9pt}\color{codegreen},
  stringstyle=\fontsize{8pt}{9pt}\color{codeblue},
  frame=tb,
  otherkeywords = {self},
}
\begin{figure}[!t]
\begin{lstlisting}[language=python]
def matrix_nms(scores, masks, method='gauss', sigma=0.5):
    # scores: mask scores in descending order (N)
    # masks: binary masks (NxHxW)
    # method: 'linear' or 'gauss'
    # sigma: std in gaussian method

    # reshape for computation: Nx(HW)
    masks = masks.reshape(N, HxW)
    # pre-compute the IoU matrix: NxN
    intersection = mm(masks, masks.T)
    areas = masks.sum(dim=1).expand(N, N)
    union = areas + areas.T - intersection
    ious = (intersection / union).triu(diagonal=1)

    # max IoU for each: NxN
    ious_cmax = ious.max(0)
    ious_cmax = ious_cmax.expand(N, N).T
    # Matrix NMS, Equation (4): NxN
    if method == 'gauss':     # gaussian
        decay = exp(-(ious^2 - ious_cmax^2) / sigma)
    else:                     # linear
        decay = (1 - ious) / (1 - ious_cmax)
    # decay factor: N
    decay = decay.min(dim=0)
    return scores * decay
\end{lstlisting}
\vspace{-1em}
\caption{Python code of Matrix NMS. \texttt{mm}: matrix multiplication; \texttt{T}: transpose; \texttt{triu}: upper triangular part.}
\label{fig:matrix_nms}
\end{figure}

\section{Experiments}

To evaluate the proposed methods, we conduct experiments on three basic tasks, instance segmentation, object detection, and panoptic segmentation on MS COCO~\cite{coco}.
Specifically, we report the results of a few variants of SOLO on COCO instance segmentation benchmark and adopt our best variant SOLOv2 in other experiments for its simplicity and high performance\footnote{The default SOLOv2 has similar performance with the dynamic version while being conceptually simpler.}.
We also present experimental results on Cityscapes~\cite{cityscapes} and LVIS~\cite{lvis2019}.
Unlike COCO, Cityscapes specifically focuses on urban street scenes.
LVIS has more than 1K categories and thus is considerably more challenging compared with COCO.
We conduct extensive ablation experiments, as well as intuitive visualization, to show how each component contributes and how SOLO framework works.

\begin{table}[tb]
\centering
\caption{Instance segmentation mask AP (\%) on COCO \texttt{test}-\texttt{dev}. All entries are \textit{single-model} results. Mask R-CNN$ ^*$ is our improved version with scale augmentation and longer training time ($6\times$).
`dcn' means deformable convolutions used. SOLOv2$^+$ is our improved version with auxiliary semantic loss and denser grid, as detailed in Section~\ref{sec:other_impr}.
}
\resizebox{.51\textwidth}{!}{%
\begin{tabular}{ r  l cccccc}
\toprule
\footnotesize
& backbone &AP & AP$_{50}$ & AP$_{75}$&AP$_{S}$ & AP$_{M}$ & AP$_{L}$\\
\midrule
\emph{box-based:} &&&&&&&\\
Mask R-CNN~\cite{maskrcnn}        &R101-FPN &35.7  &58.0   &37.8   &15.5   &38.1   &52.4 \\
Mask R-CNN$^*$                    &R101-FPN &37.8  &59.8   &40.7   &20.5   &40.4   &49.3 \\
MaskLab+~\cite{masklab}           &R101-C4  &37.3  & 59.8   & 39.6   & 16.9   & 39.9   &53.5 \\
TensorMask~\cite{Chen_2019_ICCV}  &R101-FPN &37.1  &59.3   &39.4   &17.4   &39.1   &51.6 \\
YOLACT~\cite{yolact}              &R101-FPN &31.2  &50.6   &32.8   &12.1   &33.3   &47.1 \\
MEInst~\cite{MEInst}              &R101-FPN &33.9  &56.2   &35.4   &19.8   &36.1   &42.3  \\
CenterMask~\cite{wang2020centermask} &Hourglass-104 &34.5 &56.1  &36.3   &16.3   &37.4   &48.4  \\
BlendMask~\cite{chen2020blendmask} &R101-FPN &38.4  &60.7  &41.3    &18.2   &41.5   &53.3 \\
CondInst~\cite{CondInst} &R101-FPN & 39.1 & 60.9 & 42.0 & 21.5 & 41.7 & 50.9 \\
\midrule
\emph{box-free:} &&&&&&&\\
PolarMask~\cite{polarmask}        &R101-FPN &32.1 &53.7 &33.1 &14.7 &33.8 &45.3 \\
\textbf{SOLO}   &R50-FPN  &  36.8 & 58.6 & 39.0 & 15.9 & 39.5 & 52.1\\
\textbf{SOLO}      &R101-FPN &37.8 & 59.5 & 40.4 & 16.4 & 40.6 & 54.2\\
\textbf{D-SOLO}                            &R50-FPN  &  37.4 & 58.5 & 40.0 & 16.2 & 40.3 & 52.9\\
\textbf{D-SOLO}                            &R101-FPN  &  38.4 & 59.6 & 41.1 & 16.8 & 41.5 & 54.6\\
\textbf{\method}               &R50-FPN  &  38.8 & 59.9  & 41.7  & 16.5  & 41.7  & 56.2 \\
\textbf{\method}               &R101-FPN  &  39.7 & 60.7 & 42.9 & 17.3 & 42.9 & 57.4 \\
\textbf{\method}               &R-dcn101-FPN  & 41.7 & 63.2 & 45.1 & 18.0 & 45.0& 61.6 \\
\textbf{\method}$^+$              &R-dcn101-FPN  & 42.8 & 64.6 & 46.4 & 19.9 & 45.7 & 61.9 \\
\bottomrule
\end{tabular}
}
\label{tab:sota}
\end{table}

\subsection{Main Results}
\subsubsection{MS COCO Instance Segmentation}
For our main results, we report MS COCO mask AP on the
\texttt{test}-\texttt{dev} split, which has no public labels and is evaluated on the evaluation server.

\myparagraph{Training details.}
SOLO is trained with stochastic gradient descent (SGD). We use synchronized SGD over 8 GPUs with a total of 16 images per mini-batch.
Unless otherwise specified, all models are trained for 36 epochs (\ie, 3$ \times$) with an initial learning rate of $0.01$, which is then
divided by 10 at 27th and again at 33th epoch.
Weight decay of $0.0001$ and momentum of $ 0.9$ are used. All models are initialized from ImageNet~\cite{russakovsky2015imagenet} pre-trained weights.
We use scale jittering, where the shorter image side is randomly sampled from 640 to 800 pixels.

We compare our methods to the state-of-the-art methods in instance segmentation on MS COCO \texttt{test}-\texttt{dev} in Table~\ref{tab:sota}. Our SOLOv2 with ResNet-101 achieves a mask AP of 39.7\%, which is much better than
other state-of-the-art instance segmentation methods.
Our method shows its superiority especially on large objects (\eg, ~+5.0\% AP$_{L}$ than Mask R-CNN).

\begin{figure}[t!]
    \centering
    \subfigure[Accuracy vs.\  Speed]{
        \includegraphics[width=0.438\textwidth]{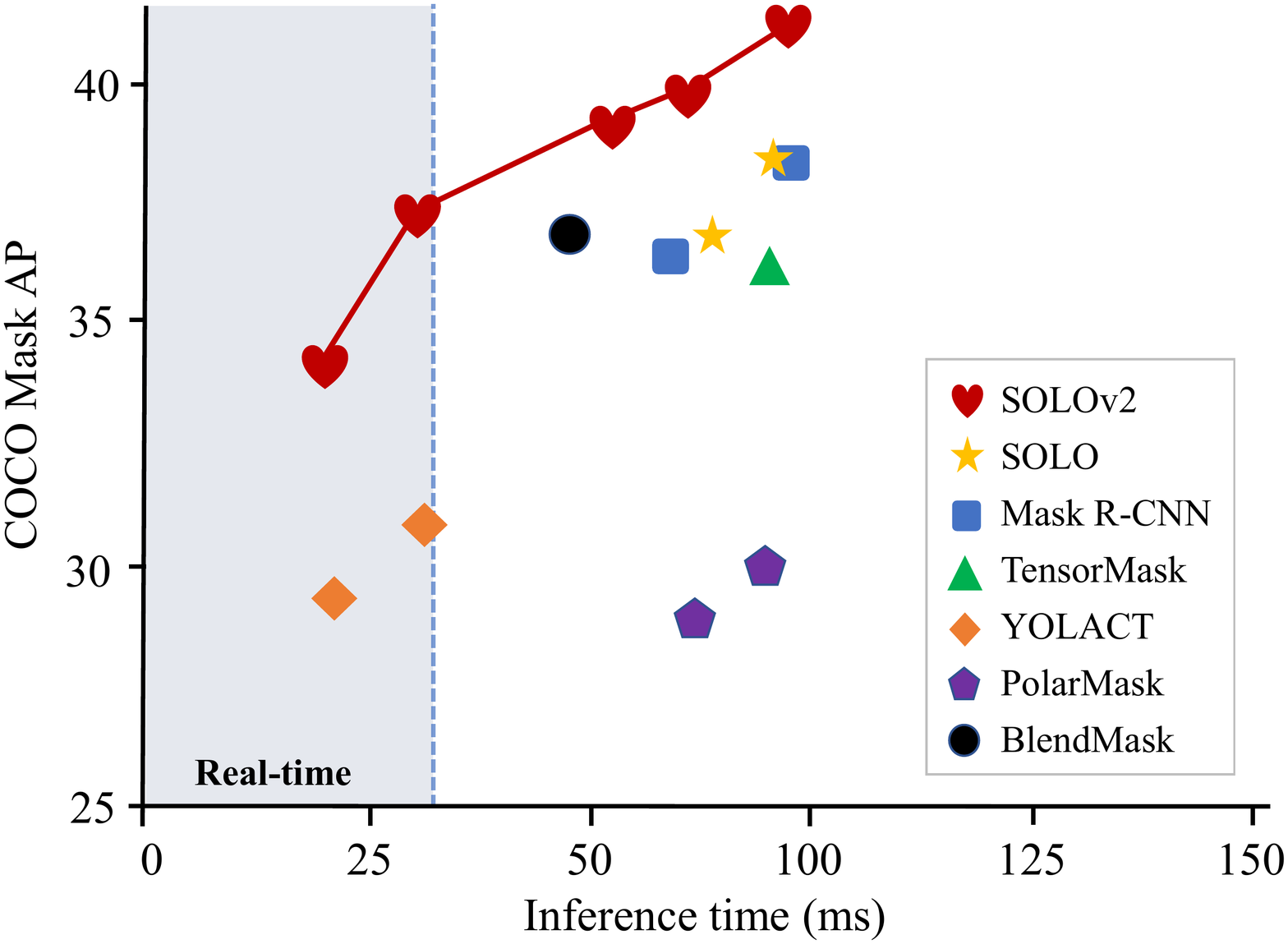}
        \label{fig:speed_vs_acc}
    }
    \subfigure[Segmentation Detail Comparison]{
        \includegraphics[width=0.46\textwidth]{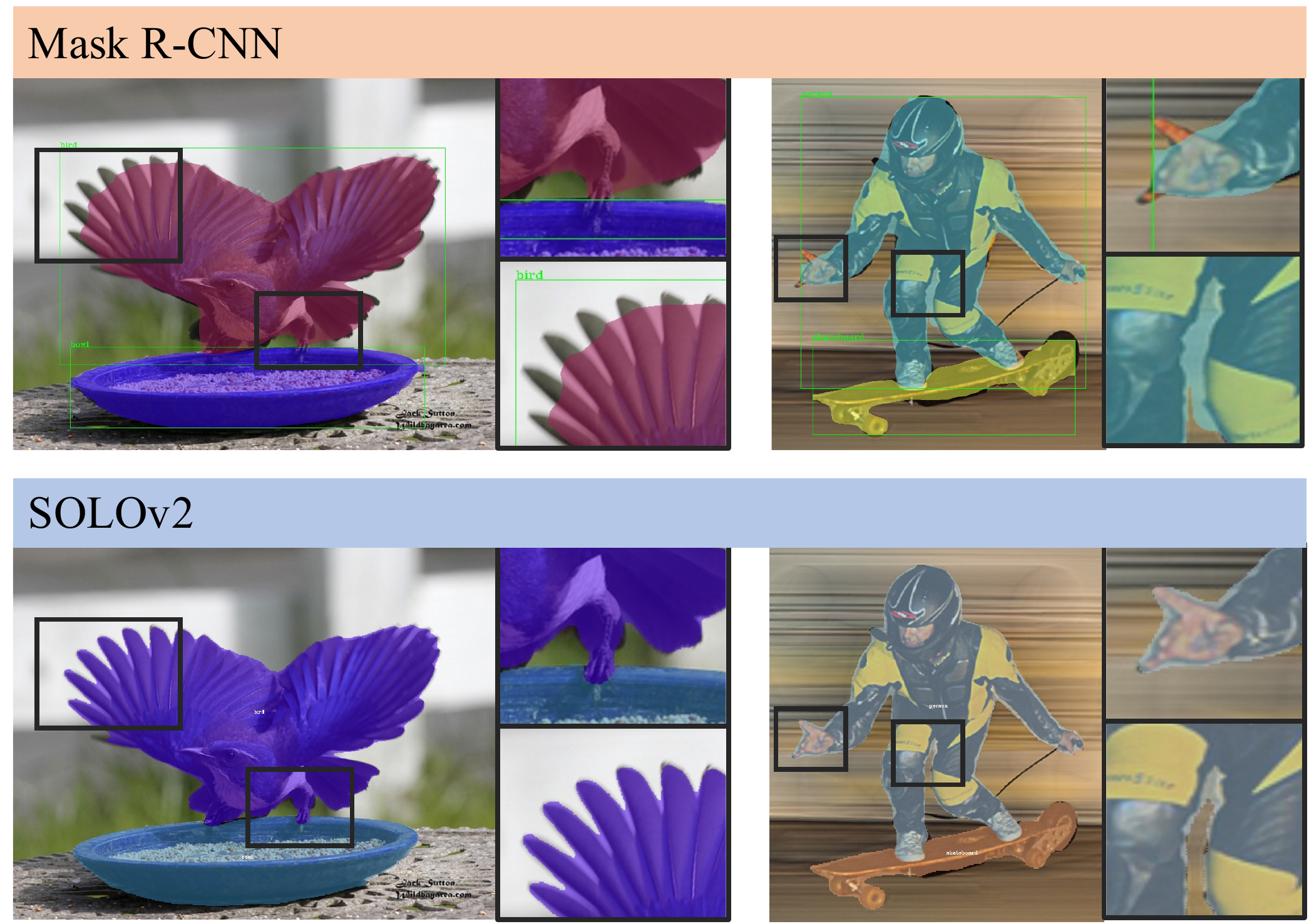}
        \label{fig:detail_solov2_vs_maskrcnn}
    }
\caption{Comparison of instance segmentation performance of SOLO and other methods on the    COCO \texttt{test}-\texttt{dev}.
    (a) The proposed method outperforms a range of state-of-the-art algorithms. All methods are evaluated using one Tesla V100 GPU.
   (b) SOLOv2 obtains higher-quality masks compared with Mask R-CNN. Mask R-CNN's mask head is typically restricted to $28 \times 28$ resolution, leading to inferior prediction at object boundaries.}
\label{fig:performance}
\end{figure}

We also provide the speed-accuracy trade-off on COCO to compare with some dominant instance segmentation algorithms (Figure~\ref{fig:performance}\textcolor{red}{(a)}).
We show our models with ResNet-50, ResNet-101, ResNet-DCN-101 and two light-weight versions described in Section~\ref{subsubsec:ablation2}.
The proposed \method outperforms a range of state-of-the-art algorithms, both in accuracy and speed.  The running time is tested on our local machine, with a single V100 GPU.
We download the code and pre-trained models to test inference time for each model on the same machine.
Further, as described in Figure~\ref{fig:performance}\textcolor{red}{(b)}, \method predicts much finer masks than Mask R-CNN.

\subsubsection{Instance Segmentation on LVIS}

LVIS~\cite{lvis2019} is a recently proposed dataset for long-tail object segmentation, which has more than 1000 object categories. In LVIS, each object instance is segmented with a
high-quality mask that surpasses the annotation quality of
the relevant COCO dataset.

Table~\ref{table:lvis}
reports the performances on the
rare (1$\sim$10 images), common (11$\sim$100), and frequent ($>$ 100) subsets, as well as the overall AP.
Both the reported Mask R-CNN and \method use data resampling training strategy, following~\cite{lvis2019}.
Our \method outperforms the baseline method by about 1\% AP. For large-size  objects (AP$_L$), our \method  achieves 6.7\%  AP improvement, which is consistent with the results on the COCO dataset.

\begin{table*}[tb]
\centering
\caption{Instance segmentation results on the
{LVIS}v0.5 validation dataset. $*$ means re-implementation.}
 \footnotesize
\begin{tabular}{ r  lccccccc}
\toprule
  & backbone & AP$_r$ & AP$_c$ & AP$_f$ & AP$_S$ & AP$_M$ & AP$_L$&AP \\
\midrule
Mask R-CNN~\cite{lvis2019}& Res-50-FPN     & 14.5 & 24.3 & 28.4 & - & - & - & 24.4 \\
Mask R-CNN$^*$-3\X & Res-50-FPN               & 12.1 & 25.8 & 28.1 & 18.7 & 31.2 & 38.2 & 24.6 \\
\textbf{\method}  & Res-50-FPN      & 13.4 & 26.6 & 28.9 & 15.9 & 34.6 & 44.9 & 25.5 \\
\textbf{\method}  & Res-101-FPN     & 16.3 & 27.6 & 30.1 & 16.8 & 35.8 & 47.0 & 26.8 \\
\bottomrule
\end{tabular}
\label{table:lvis}
\end{table*}

\subsubsection{ Instance Segmentation on Cityscapes}
Cityscapes~\cite{cityscapes} is another popular instance segmentation benchmark which only focuses on urban street scenes.
The size of images in Cityscapes is much larger than that in MS COCO.
We double the grid numbers for the experiments on Cityscapes.
As shown in Table~\ref{table:cityscapes}, \method outperforms Mask R-CNN by 0.9\% AP, which demonstrates the effectiveness of our method.

\begin{table}[tb]
\centering
\caption{Instance segmentation results on the
Cityscapes \texttt{val}. $*$ means re-implementation.}
 \footnotesize
\begin{tabular}{ r  lc}
\toprule
  & backbone & AP \\
\midrule
Mask R-CNN~\cite{maskrcnn}& Res-50-FPN     &  36.4 \\
Mask R-CNN$^*$ & Res-50-FPN               & 36.5 \\
\textbf{\method}  & Res-50-FPN      & 37.4 \\
\textbf{\method}  & Res-101-FPN     & 38.0 \\
\bottomrule
\end{tabular}
\label{table:cityscapes}
\end{table}

\subsection{Ablation Study}

\subsubsection{SOLO Ablation}

\myparagraph{Grid number.}
We compare the impacts of grid number on the performance with single output feature map as shown in Table~\ref{tab:numgrid}\textcolor{red}{(a)}. The feature is generated by merging C3, C4, and C5 outputs in ResNet (stride: 8). To our surprise, $S = 12$
can already achieve
27.2\%  AP on the challenging MS COCO dataset. SOLO achieves 29\%  AP when improving the grid number to 24. This results indicate that our single-scale SOLO
can be applicable to some scenarios where object scales do not vary much.

\myparagraph{Multi-level Prediction.}
From Table~\ref{tab:numgrid}\textcolor{red}{(a)} we can see that our single-scale SOLO
could already get 29.0\% AP on MS COCO dataset. In this ablation, we show that the performance could be further improved via multi-level prediction using FPN~\cite{fpn}.
We use five pyramids to segment objects of different scales (details in supplementary).  Scales of ground-truth masks are explicitly used to assign them to the levels of the pyramid.
From P2 to P6, the corresponding grid numbers are $[40, 36, 24, 16, 12]$ respectively.
Based on our multi-level prediction, we further achieve 35.8\% AP. As expected, the performance over all the metrics has been largely improved.

%

\myparagraph{CoordConv.} Another important component that facilitates our SOLO paradigm is the \textit{spatially variant} convolution (CoordConv~\cite{coordconv}). As shown in Table~\ref{tab:coordconv}\textcolor{red}{(b)}, the
standard convolution can already have spatial variant property to some extent, which is in accordance with the observation in~\cite{coordconv}.
As also revealed in~\cite{Islam2020How}, CNNs can implicitly learn the absolute position information from the commonly used zero-padding operation.
However, the implicitly learned position information is coarse and inaccurate.
When making the convolution access to its own input coordinates through
concatenating extra coordinate channels,
our method enjoys 3.6\% absolute AP gains.
Two or more CoordConvs do not bring
noticeable
improvement.
It suggests that a single CoordConv already enables the predictions to be well spatially variant/position sensitive.

\myparagraph{Loss function.}
Table \ref{tab:loss}\textcolor{red}{(c)} compares different loss functions for our mask optimization branch. The methods include conventional Binary Cross Entropy (BCE), Focal Loss (FL), and Dice Loss (DL).
To obtain improved performance, for Binary Cross Entropy, we set a mask loss weight of 10 and a pixel weight of 2 for positive samples.
The mask loss weight of Focal Loss is set to 20.
As shown, the Focal Loss works much better than ordinary Binary Cross Entropy loss.
It is because that the majority of pixels of an instance mask are in background, and the Focal Loss is
designed to mitigate
the sample imbalance problem by decreasing the loss of well-classified samples.
However, the Dice Loss achieves the best results without the need of manually adjusting the loss hyper-parameters.
Dice Loss
views the pixels as a whole object and could establish
the right balance between foreground and background pixels automatically.
Note that with carefully tuning the balance hyper-parameters and introducing
other training tricks,  the results of Binary Cross Entropy
and Focal Loss may be considerably improved. However, the point here is that with
the
Dice Loss, training  typically becomes much more stable and more likely to attain good results without using much heuristics.
To make a fair comparison, we also show the results of Mask R-CNN with Dice loss.
 In Mask R-CNN, replacing the original BCE loss with Dice loss gives a $-0.9$\% AP drop.

\myparagraph{Alignment in the category branch.}
In the category prediction branch, we must match the convolutional features with spatial size $H$$\times$$W$ to $S$$\times$$S$. Here, we compare three common implementations: interpolation, adaptive-pool, and region-grid-interpolation. (a) Interpolation: directly bilinear interpolating to the target grid size; (b) Adaptive-pool: applying a 2D adaptive max-pool over $H$$\times$$W$ to $S$$\times$$S$; (c) Region-grid-interpolation: for each grid cell, we use bilinear interpolation conditioned on dense sample points, and aggregate the results with average.
From our observation, there is no noticeable performance gap between these variants ($\pm$ 0.1\% AP), indicating that the alignment process
does not have a significant impact on the final accuracy.

\myparagraph{Different head depth.}
In SOLO, instance segmentation is formulated as a pixel-to-pixel task, and we exploit the spatial layout of masks by using an FCN. In Table~\ref{tab:depth_ap}\textcolor{red}{(d)}, we compare different head depth used in our work. Changing the head depth from 4 to 7 gives 1.2\% AP gains.  The results show that when the depth grows beyond 7, the performance becomes stable. In this paper, we  use depth being 7 in other experiments of vanilla SOLO.

\myparagraph{Vanilla vs.\ Decoupled.}
As shown in Table~\ref{tab:sota}, decoupled SOLO performs slightly better than vanilla SOLO (37.4\% vs.\ 36.8\% AP).
Here, we qualitatively and quantitatively analyze why decoupled SOLO has better performance.
For mask prediction, vanilla SOLO defines $S^2$ location categories.
Give an object in the input image, the model needs to figure out which location category this object belongs to.
When $S$ is large, \eg, 40, it is challenging to distinguish $S^2$  location categories.
By contrast, decoupled SOLO decouples the  $S^2$ location categories into $2S$ location categories, \ie, $S$ horizontal  categories and  $S$ vertical  categories.
For each pixel of an input object, the model only needs to classify it into $2S$ categories, which is considerably simpler than that in vanilla SOLO.

To show a more intuitive comparison, we plot the training curves of vanilla SOLO and decoupled SOLO.
As shown in
Figure 
\ref{fig:loss_curves}
in the supplementary,
the loss curves of category prediction are almost the same.
By contrast, the decoupled SOLO shows better convergence in mask loss compared to vanilla SOLO.
The comparison of the mask loss curve indicates the superiority of the decoupled mask representation.
\begin{table}[!t]
\caption{Ablation experiments for vanilla SOLO. All models are trained on MS COCO \texttt{train2017} and tested on \texttt{val2017} unless noted.}
\vskip 0.05 in
\begin{minipage}{0.99\linewidth}
\begingroup
{ (a) The impact of \textbf{grid number and FPN}. FPN significantly improves the performance
 thanks to its ability to deal with varying sizes of objects.}
\endgroup
\vskip 0.05 in
\centering
\scalebox{0.9}{
\begin{tabular}{ccccccc}
\toprule
        grid number &AP & AP$_{50}$ & AP$_{75}$ & AP$_{S}$ & AP$_{M}$ & AP$_{L}$\\
        \midrule
        12&  27.2 & 44.9  & 27.6  & 8.7  & 27.6  & 44.5   \\
        24&  29.0  & 47.3  & 29.9  & 10.0  & 30.1  & 45.8    \\
        36&   28.6  & 46.3  & 29.7  & 9.5  & 29.5  & 45.2   \\
        Pyramid & 35.8  & 57.1  & 37.8  & 15.0  & 38.7  & 53.6  \\
    \bottomrule
    \end{tabular}}
\label{tab:numgrid}
\end{minipage}
\vskip 0.15 in
\begin{minipage}{.99\linewidth}
\begingroup
{ (b) \textbf{Conv vs.\  CoordConv.}
    CoordConv can considerably improve AP upon standard convolution.
    Two or more layers of CoordConv are not necessary. }
\endgroup
\vskip 0.05 in
\centering
\scalebox{0.92}{
\begin{tabular}{ccccccc}
\toprule
        \#CoordConv &AP & AP$_{50}$ & AP$_{75}$ & AP$_{S}$ & AP$_{M}$ & AP$_{L}$\\
        \midrule
        0&  32.2 & 52.6 & 33.7  & 11.5  & 34.3   & 51.6    \\
        1&  35.8  & 57.1  & 37.8  & 15.0  & 38.7  & 53.6   \\
        2&  35.7 & 57.0 & 37.7 & 14.9 & 38.7 & 53.3   \\
        3&  35.8 & 57.4 & 37.7 & 15.7 & 39.0 & 53.0   \\
        \bottomrule
    \end{tabular}
}
\label{tab:coordconv}
\end{minipage}
\vskip 0.15 in
\begin{minipage}{.99\linewidth}
\begingroup
{ (c) \textbf{Different loss functions} may be employed in the mask branch. The Dice
loss (DL) leads to best AP and is more stable to train.}
\endgroup
\vskip 0.05 in
\centering
\scalebox{0.9}{
\begin{tabular}{ccccccc}
\toprule
        mask loss &AP & AP$_{50}$ & AP$_{75}$ & AP$_{S}$ & AP$_{M}$ & AP$_{L}$\\
        \midrule
        BCE& 30.0 & 50.4  & 31.0  & 10.1  & 32.5  & 47.7   \\
        FL& 31.6 & 51.1 & 33.3 & 9.9 & 34.9 & 49.8  \\
        DL&  35.8  & 57.1  & 37.8  & 15.0  & 38.7  & 53.6   \\
         \bottomrule
    \end{tabular}
}
\label{tab:loss}
\end{minipage}
\vskip 0.15 in
\begin{minipage}{.99\linewidth}
\begingroup
{ (d) \textbf{Head depth.} We  use depth being 7 in other vanilla SOLO experiments.}
\endgroup
\vskip 0.05 in
\centering
\scalebox{0.92}{
\begin{tabular}{cccccc}
\toprule
        head depth & 4 & 5 & 6 & 7 & 8 \\
        \midrule
        AP & ~34.6~ & ~35.2~ & ~35.5~ & ~35.8~ & ~35.8~  \\
        \bottomrule
    \end{tabular}
}
\label{tab:depth_ap}
\end{minipage}
%
\label{table:ablation1}
\end{table}

\subsubsection{SOLOv2 Ablation}
\label{subsubsec:ablation2}

\myparagraph{Kernel shape.}
We consider the kernel shape from two aspects: number of input channels and kernel size.
The comparisons are shown in Table~\ref{tab:kernel_shape}\textcolor{red}{(a)}.
$1\times1$ conv.\  shows equivalent performance to $3\times3$ conv.
Changing the number of input channels from 128 to 256
attains
0.4\%  AP gains.  When it grows beyond 256, the performance becomes stable. In this work, we  set the number of input channels to be 256 in all other experiments.

\myparagraph{Effectiveness of coordinates.}
Since our method segments objects by locations, or specifically, learns the object segmenters by locations, the position information is very important.
For example, if the mask kernel branch is unaware of the positions, the objects with the same appearance may have the same predicted kernel, leading to the same output mask.
On the other hand, if the mask feature branch is unaware of the position information, it would not know how to assign the pixels to different feature channels in the order that matches the mask kernel.
As shown in Table~\ref{tab:coordinates}\textcolor{red}{(b)},
the model achieves 36.3\%  AP without explicit coordinates input.
The results are
reasonably good
because that CNNs can implicitly learn the absolute position information from the commonly used zero-padding operation, as revealed in~\cite{Islam2020How}.
The pyramid zero-paddings in our mask feature branch should have contributed considerably.
However, the implicitly learned position information is coarse and inaccurate.
When making the convolution access to its own input coordinates through
concatenating extra coordinate channels,
our method enjoys 1.5\% absolute AP gains.

\myparagraph{Unified mask feature representation.}
For mask feature learning, we have two options: to learn the feature in the head separately for each FPN level, or to construct a unified representation.
For the former one, we implement as SOLO and use
seven
$3\times3$ convolutions  to predict the mask features.
For the latter one, we fuse the FPN's features in a simple way and obtain the unified mask representations. The detailed implementation is provided in supplementary material.
We compare these two modes in Table~\ref{tab:mask_fea}\textcolor{red}{(c)}.
As shown, the unified representation achieves better results, especially for the medium and large objects.
This is easy to understand:
In a separate way, the large-size objects are assigned to
high-level feature maps of low spatial resolutions, leading to
coarse boundary prediction.

\iftrue
\myparagraph{Dynamic vs.\ Decoupled.}
 The dynamic head and decoupled head both serve as the efficient varieties of the SOLO head.
 We compare the results in Table~\ref{tab:dynamic_vs_decouple}\textcolor{red}{(d)}.
 All the settings are the same except the head type, which means that for the dynamic head we use the separate features as above.
 The  dynamic head
 achieves
 0.7\% AP better than the decoupled head.
We believe that the gains
have
come from the dynamic scheme which learns the kernel weights dynamically,
conditioned on the input.
\fi

\myparagraph{Matrix NMS.}  Our Matrix NMS can be implemented totally in parallel. Table~\ref{tab:nms}\textcolor{red}{(e)}
presents
the speed and accuracy comparison of Hard-NMS, Soft-NMS, Fast NMS and our Matrix NMS. Since all methods need to compute the IoU matrix, we pre-compute the IoU matrix in advance for fair comparison. The speed reported here is
that of the
NMS process alone, excluding computing IoU matrices.
Hard-NMS and Soft-NMS are widely used in current object detection and segmentation models. Unfortunately, both methods are
recursive
and spend much time budget (\eg,  22 ms). Our Matrix NMS only needs
less than
1
ms and is almost cost free.  Here we also show the performance of Fast NMS, which
also
utilizes matrix operations but with performance penalty. To conclude, our Matrix NMS shows its advantages on both speed and accuracy.

\myparagraph{Real-time setting.}
We design two light-weight models for different purposes.
1) \texttt{Speed priority}, the number of
convolution layers in the prediction head is reduced to two and the input shorter side is 448.
2) \texttt{Accuracy priority}, the number of
convolution layers in the prediction head is reduced to three and the input shorter side is 512. Moreover, deformable convolution~\cite{dai2017deformable} is used in the backbone and the last layer of prediction head.
We train both models with the $3\times$ schedule, with the shorter side randomly sampled from [352, 512].
Results are shown in Table~\ref{tab:real_time}\textcolor{red}{(f)}.
\method can not only push state-of-the-art, but has also been ready for real-time applications.

\begin{figure}[b]
\centering
\includegraphics[width=0.334\textwidth]{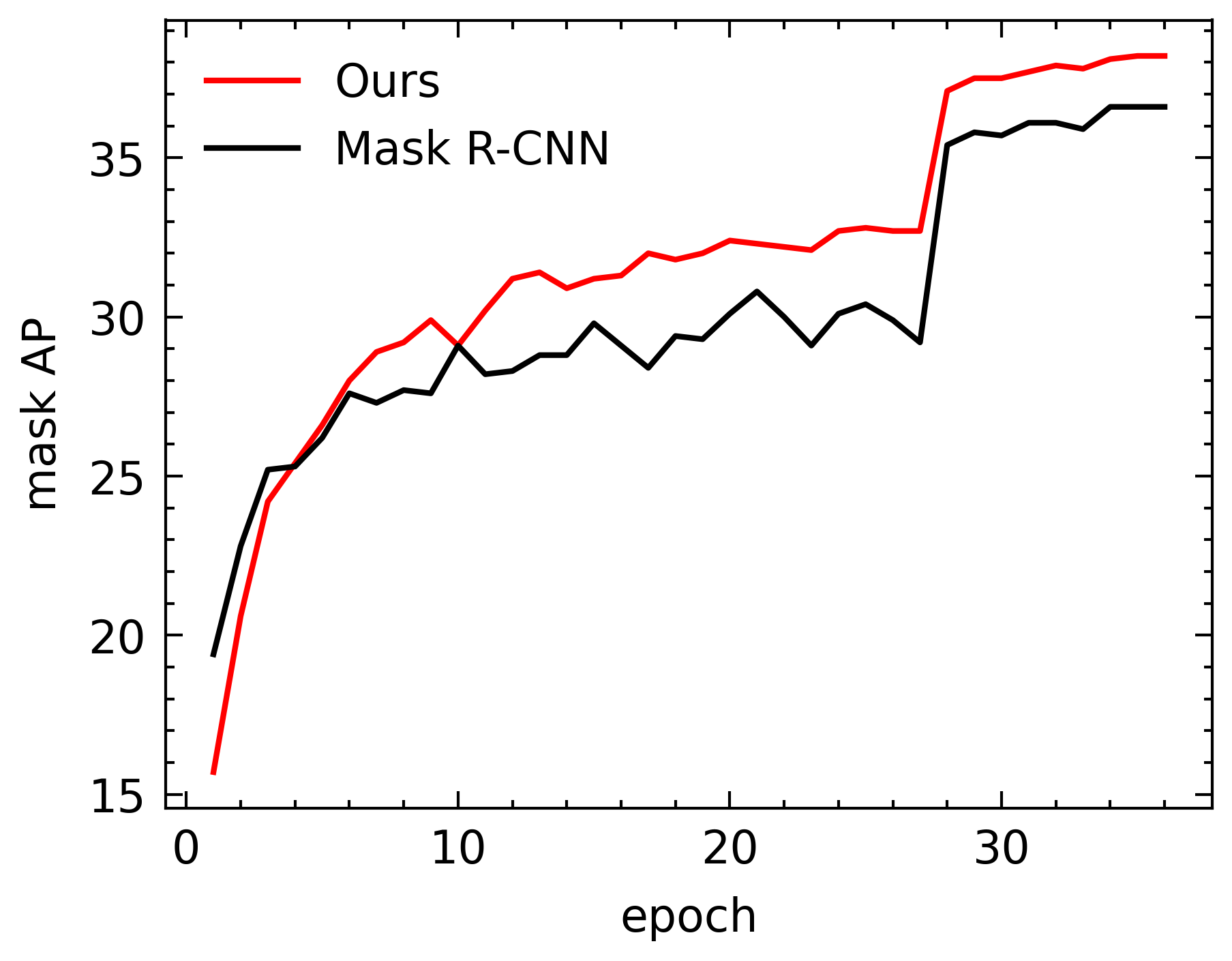}
\caption{Instance segmentation results %
of
different numbers of training epochs. Ours is the SOLOv2 model with denser grid. The models are
trained on MS COCO \texttt{train2017} with %
36 epochs,
and tested on \texttt{val2017}.}
\label{fig:ap_curve}
\end{figure}

\myparagraph{Training schedule.}
We report the results of different training schedules in Table~\ref{tab:ablation_schedule}\textcolor{red}{(g)}, including 12 epochs using single-scale training (1$\times$) and  36 epochs with multi-scale training (3$\times$).
To further show how training time affects our method, we have plotted the AP curve and compared it with the previous dominant Mask R-CNN.
As shown in Figure~\ref{fig:ap_curve}, Mask R-CNN shows a slight advantage in the first 5 epochs.
As the training time increases, our method consistently outperforms it by a large margin.

\begin{table}[!t]
\caption{Ablation experiments for \method. All models are trained on MS COCO \texttt{train2017} and tested on \texttt{val2017} unless noted.}
\vskip 0.05 in
\begin{minipage}{0.99\linewidth}
\begingroup
{ (a) \textbf{Kernel shape.}  The performance is stable when the shape goes beyond $1\times1\times256$.}
\endgroup
\vskip 0.05 in
\centering
\scalebox{0.92}{
\begin{tabular}{ r cccccc}
\toprule
        Kernel shape &AP & AP$_{50}$ & AP$_{75}$ & AP$_{S}$ & AP$_{M}$ & AP$_{L}$\\
        \midrule
        $3\times3\times64$ &  37.4 & 58.0 & 39.9 & 15.6 & 40.8 & 56.9 \\
        $1\times1\times64$&  37.4  & 58.1  & 40.1 & 15.5 & 41.1 & 56.3\\
        $1\times1\times128$&  37.4 & 58.1 & 40.2 & 15.8 & 41.1 & 56.6   \\
        $1\times1\times256$&  37.8 & 58.5 & 40.4  & 15.6 & 41.3 & 56.8 \\
        $1\times1\times512$&  37.7 & 58.3 & 40.4 & 15.4 & 41.5 & 56.6   \\
    \bottomrule
    \end{tabular}}
\label{tab:kernel_shape}
\end{minipage}
\vskip 0.15 in
\begin{minipage}{.99\linewidth}
\begingroup
{ (b) \textbf{Explicit coordinates.} Precise coordinates input can considerably improve the results.}
\endgroup
\vskip 0.05 in
\centering
\scalebox{0.92}{
\begin{tabular}{p{.7cm}<{\centering} p{1cm}<{\centering} cccccc}
\toprule
        Kernel & Feature &AP & AP$_{50}$ & AP$_{75}$ & AP$_{S}$ & AP$_{M}$ & AP$_{L}$ \\
        \midrule
         & & 36.3 & 57.4 & 38.6  & 15.6 & 39.8 & 54.7   \\
         \cmark & &  36.3  & 57.3  & 38.5  & 15.1 & 40.0 & 54.1 \\
        & \cmark &  37.1 & 58.0 & 39.4 & 15.2 & 40.5 & 55.9  \\
        \cmark & \cmark & 37.8 & 58.5 & 40.4 & 15.6 & 41.3 & 56.8  \\
        \bottomrule
    \end{tabular}
}
\label{tab:coordinates}
\end{minipage}
\vskip 0.15 in
\begin{minipage}{.99\linewidth}
\begingroup
{ (c) \textbf{Mask feature representation.} We compare the separate mask feature representation in parallel heads and the unified representation.}
\endgroup
\vskip 0.05 in
\centering
\scalebox{0.92}{
\begin{tabular}{ccccccc}
\toprule
        Mask Feature &AP & AP$_{50}$ & AP$_{75}$ & AP$_{S}$ & AP$_{M}$ & AP$_{L}$  \\
        \midrule
        Separate  &  37.3 & 58.2 & 40.0 & 15.7 & 40.8 & 55.5\\
        Unified &  37.8 & 58.5 & 40.4 & 15.6 & 41.3 & 56.8  \\
        \bottomrule
    \end{tabular}
}
\label{tab:mask_fea}
\end{minipage}
\vskip 0.15 in
\begin{minipage}{.99\linewidth}
\begingroup
{ (d) \textbf{Dynamic head vs.\  Decoupled head.} Dynamic$^*$ indicates that the dynamic head here is applied on separate features outputted by 7 convs in parallel. }
\endgroup
\vskip 0.05 in
\centering
\scalebox{0.92}{
\begin{tabular}{ccccccc}
\toprule
        Head type &AP & AP$_{50}$ & AP$_{75}$ & AP$_{S}$ & AP$_{M}$ & AP$_{L}$\\
        \midrule
        Decoupled  &  36.6 & 57.2 & 39.0 & \textbf{16.0} & 40.3 & 53.5  \\
        Dynamic$^*$ &  \textbf{37.3} & \textbf{58.2} & \textbf{40.0} & 15.7 & \textbf{40.8} & \textbf{55.5} \\
        \bottomrule
    \end{tabular}
}
\label{tab:dynamic_vs_decouple}
\end{minipage}
\vskip 0.15 in
\begin{minipage}{.99\linewidth}
\begingroup
{ (e) \textbf{Matrix NMS.} Matrix NMS outperforms other methods in both speed and accuracy.}
\endgroup
\vskip 0.05 in
\centering
\scalebox{0.92}{
\begin{tabular}{ r ccc}
\toprule
        Method & Iter? & Time (ms) & AP \\
        \midrule
         Hard-NMS & \cmark & 9 & 36.3    \\
         Soft-NMS & \cmark  & 22   & 36.5   \\
          Fast NMS & \xmark & $\textless1$ &  36.2  \\
         Matrix NMS & \xmark & $\textless1$ & 36.6   \\
         \bottomrule
    \end{tabular}
}
\label{tab:nms}
\end{minipage}
\vskip 0.15 in
\begin{minipage}{.99\linewidth}
\begingroup
{ (f) \textbf{Real-time \method}. The speed is reported on a single V100 GPU by averaging 5 runs (on COCO \texttt{test-dev}). }
\endgroup
\vskip 0.05 in
\centering
\scalebox{0.88}{
\begin{tabular}{c c c c c c c c}
\toprule
    Model  & AP & AP$_{50}$ & AP$_{75}$ & AP$_{S}$ & AP$_{M}$ & AP$_{L}$ & fps \\
        \midrule
         \method-448  & 34.0 & 54.0 & 36.1 & 10.3 & 36.3 & 54.4 & 46.5   \\
         \method-512  & 37.1 & 57.7 & 39.7 & 12.9 & 40.0 & 57.4 & 31.3 \\
         \bottomrule
    \end{tabular}
}
\label{tab:real_time}
\end{minipage}

\vskip 0.15 in
\begin{minipage}{.99\linewidth}
\begingroup
{ (g) \textbf{Training schedule.} 1$\times$ means 12 epochs using single-scale training. 3$\times$ means 36 epochs with multi-scale training.}
\endgroup
\vskip 0.05 in
\centering
\scalebox{0.92}{
\begin{tabular}{ccccccc}
\toprule
        Schedule &AP & AP$_{50}$ & AP$_{75}$ & AP$_{S}$ & AP$_{M}$ & AP$_{L}$  \\
        \midrule
        1$\times$ &  34.8 & 54.8 & 36.8  & 13.1 & 38.0 & 53.8 \\
        3$\times$ &  37.8 & 58.5 & 40.4 & 15.6 & 41.3 & 56.8   \\
        \bottomrule
    \end{tabular}
}
\label{tab:ablation_schedule}
\end{minipage}

\label{table:ablation2}
\end{table}

\subsection{Other Improvements}
\label{sec:other_impr}
In this section, we further explore some potential techniques to improve our method, including auxiliary semantic loss, denser grid, decoupled mask representation, and multi-scale testing.
The ablation experiments are conducted on our SOLOv2 model with ResNet-101 backbone. All models are trained for 36 epochs (\ie, 3$\times$ schedule) on COCO \texttt{train2017}, and evaluated on \texttt{val2017}.
With auxiliary semantic loss and denser grid, we boost the results of our best single model from 41.7\% AP to 42.8\% AP on MS COCO \texttt{test-dev}, as reported in Table~\ref{tab:sota}.

\myparagraph{Auxiliary semantic loss.}
Inspired by~\cite{yolact, chen2020blendmask}, we apply the auxiliary semantic loss during the training.
Specifically, a semantic segmentation branch is added to the FPN features P2, which consists of three $3\times3$ convs.
The number of output channels is $C$, which is the number of object classes, \ie, 80 for COCO dataset.
The labels of semantic segmentation are generated from the instance segmentation annotations by merging the object masks of the same class.
As shown in Table~\ref{tab:aux}\textcolor{red}{(a)}, the auxiliary semantic loss brings 0.3\% AP gains.

\myparagraph{Denser grid.}
In our method, a pyramid of number of grid is used during training and inference.
The density of the grid decides how many objects the model can detect.
To improve the capability of our model, we increase the number of grid from $[40, 36, 24, 16, 12]$ to $[80, 64, 32, 24, 12]$.
The denser grid gives 0.7\% AP gains over the strong baseline as shown in Table~\ref{tab:denser}\textcolor{red}{(b)}.

\myparagraph{Decoupled dynamic SOLO.}
We propose a more unified algorithm by combining decoupled SOLO and dynamic SOLO.
Specifically, based on dynamic SOLO, we further decouple the mask representation inspired by decoupled SOLO.
In this way, we only need a compact mask feature $F$ with reduced output space, \eg, from
$H\times W\times 256$ to $H\times W\times 64$ for $A = 4$.
As shown in Table~\ref{table:decoupled_solov2}\textcolor{red}{(c)}, we achieve similar performance using a compact mask feature $F  \in \mathbb{ R}^{H\times W\times 64}$, compared to the  $F  \in \mathbb{ R}^{H\times W\times 256}$ in SOLOv2 without decoupled mask representation.
It should be attributed to the improved capacity of the mask representation.
The default dynamic scheme in SOLOv2 acts as the linear combination of mask feature maps.
The decoupled mask representation improves the capacity through ensembling multiple sub-masks.
It enables to maintain high performance with a compact mask feature.

\myparagraph{Multi-scale testing.}
In Table~\ref{tab:mstest}\textcolor{red}{(d)}, we report the results of multi-scale testing.
During the inference, we feed images with different scales to the model and fuse the outputs as the final predictions.
Specifically, we use three scales (the shorter size being 600, 800 and 1000) to get the corresponding outputs and apply Matrix NMS again on these outputs.

\begin{table}[!t]
\caption{Other improvements on SOLOv2 with ResNet-101 backbone. The models are trained on MS COCO \texttt{train2017} and tested on \texttt{val2017}.}
\vskip 0.05 in
\begin{minipage}{0.99\linewidth}
\begingroup
{ (a) \textbf{Auxiliary semantic loss.} The additional semantic segmentation branch improves the baseline by 0.3\% AP.}
\endgroup
\vskip 0.05 in
\centering
\scalebox{0.9}{
\begin{tabular}{ccccccc}
\toprule
        aux &AP & AP$_{50}$ & AP$_{75}$ & AP$_{S}$ & AP$_{M}$ & AP$_{L}$\\
        \midrule
         & 39.1 & 59.8 & 41.9 & 16.4 & 43.2 & 58.8  \\
        \checkmark & 39.4 & 60.1 & 42.3 & 16.4 & 43.4 & 59.6  \\
    \bottomrule
    \end{tabular}}
\label{tab:aux}
\end{minipage}
\vskip 0.15 in
\begin{minipage}{.99\linewidth}
\begingroup
{ (b) \textbf{Denser grid.} The performance on small objects (AP$_{S}$) is largely improved.}
\endgroup
\vskip 0.05 in
\centering
\scalebox{0.92}{
\begin{tabular}{ccccccc}
\toprule
        denser &AP & AP$_{50}$ & AP$_{75}$ & AP$_{S}$ & AP$_{M}$ & AP$_{L}$\\
        \midrule
         & 39.1 & 59.8 & 41.9 & 16.4 & 43.2 & 58.8  \\
        \checkmark & 39.8 & 60.7 & 42.6 & 18.6 & 43.5 & 58.2  \\
        \bottomrule
    \end{tabular}
}
\label{tab:denser}
\end{minipage}
\vskip 0.15 in
\begin{minipage}{.99\linewidth}
\begingroup
{ (c) \textbf{Decoupled mask representation.} Decoupled dynamic SOLO enables to maintain high performance with a compact mask feature, \ie,  $F  \in \mathbb{ R}^{H\times W\times 64}$.}
\endgroup
\vskip 0.05 in
\centering
\scalebox{0.92}{
\begin{tabular}{ccccccc}
\toprule
        decoupled  &AP & AP$_{50}$ & AP$_{75}$ & AP$_{S}$ & AP$_{M}$ & AP$_{L}$\\
        \midrule
          & 39.1 & 59.8 & 41.9 & 16.4 & 43.2 & 58.8  \\
        \checkmark & 39.3 & 59.8 & 42.3 & 17.1 & 43.4 & 58.7 \\
    \bottomrule
\end{tabular}
}
\label{table:decoupled_solov2}
\end{minipage}
\vskip 0.15 in
\begin{minipage}{.99\linewidth}
\begingroup
{ (d) \textbf{Multi-scale testing.} Two additional scales bring 0.4\% AP gains.}
\endgroup
\vskip 0.05 in
\centering
\scalebox{0.92}{
\begin{tabular}{ccccccc}
\toprule
        ms-test &AP & AP$_{50}$ & AP$_{75}$ & AP$_{S}$ & AP$_{M}$ & AP$_{L}$\\
        \midrule
         & 39.1 & 59.8 & 41.9 & 16.4 & 43.2 & 58.8  \\
        \checkmark & 39.5 & 60.4 &  42.4 & 17.0 & 43.4 & 59.1 \\
        \bottomrule
    \end{tabular}
}
\label{tab:mstest}
\end{minipage}
\label{table:other_impr}
\end{table}

\subsection{Qualitative Analysis}
We visualize what our method has learned
from two aspects: mask feature behavior and the final outputs after being convolved by the dynamically learned convolution kernels.

\begin{figure}[t]
\begin{center}
    \includegraphics[width=0.92\linewidth]{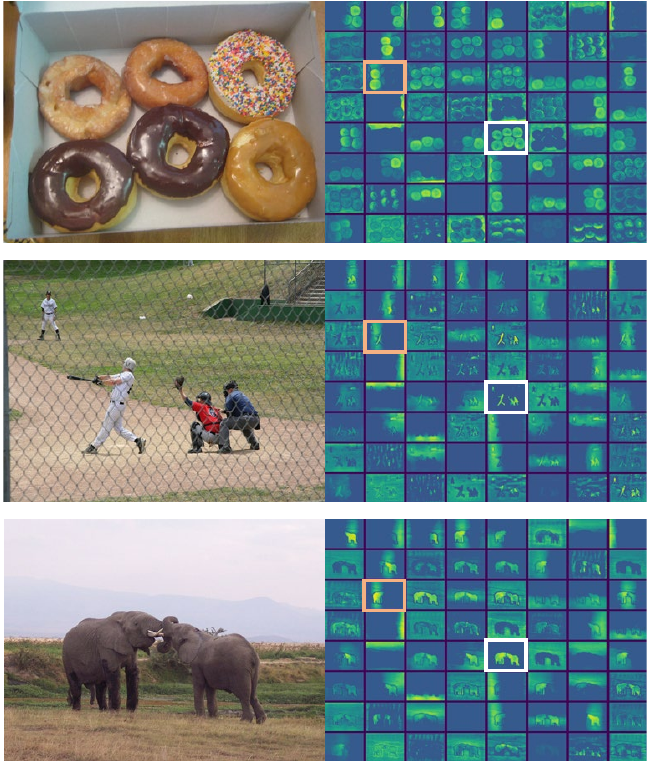}
\end{center}
\vspace{-0.1in}
   \caption{\textbf{Mask feature behavior.}
   Each plotted subfigure corresponds to one of the 64 channels of the last feature map prior to mask prediction.
   The mask features
   appear to be
   position-sensitive (orange box on the left), while a few mask features are position-agnostic and activated on all instances (white box on the right). Best viewed on screens.}
\label{fig:behavior}
\end{figure}

We visualize
the outputs of mask feature branch.
We use a model which has 64 output channels  (\ie,  $E=64$ for the last feature map prior to mask prediction) for
easy visualization.
Here  we plot each of the 64 channels (recall the channel spatial resolution is $ H \times W $) as shown in Figure~\ref{fig:behavior}.

There are two main patterns.
The first and the foremost, the mask features are position-aware.
It shows obvious behavior of scanning the objects in the image horizontally and vertically.
The other obvious pattern is that some feature maps are responsible for activating all the foreground objects, \eg, the one in white boxes.

The final outputs are shown in Figure
 \ref{fig:Vis}
in the supplementary.
Different objects are in different colors.
Our method shows promising results in diverse scenes.
It
is worth
noting
that the details at the boundaries are segmented
well, especially for large objects.

\subsection{Extension: Object Detection}

Although our instance segmentation solution removes the dependence of bounding box prediction, we are able to
produce the 4D object bounding box from each instance mask. In Table  
\ref{table:box_det}  
in the supplementary, we compare the generated box detection performance with other object detection methods on MS COCO. All models are trained on
the
{\tt train2017} subset and tested on
{\tt test-dev.}

\begin{figure}[tb]
\centering
        \includegraphics[width=0.83\linewidth]{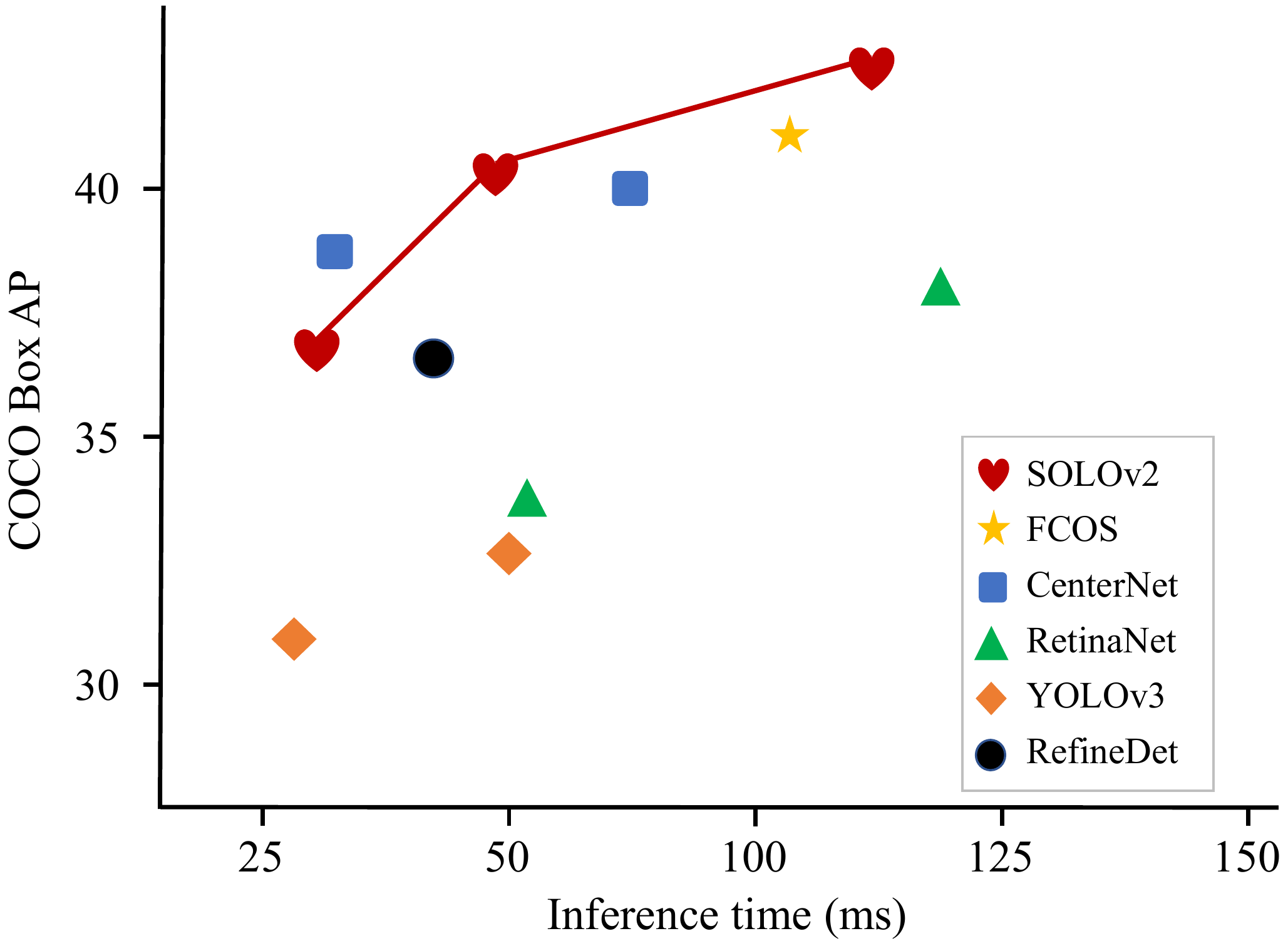}
    \caption{\method for \textbf{object detection}.
        Speed-accuracy trade-off of bounding-box detection on the COCO \texttt{test}-\texttt{dev}.
        \label{fig:box_speed_ap}
    }
\end{figure}

As shown in %
Table \ref{table:box_det}  in the supplementary,
our detection results outperform most methods, especially for objects of large sizes,
demonstrating the effectiveness of \method in bounding-box object  detection.
We also plot the speed/accuracy trade-off curve for different methods
in Figure~\ref{fig:box_speed_ap}.
We show our models with ResNet-101 and two light-weight versions described above.
The plot reveals that the bounding box performance of \method beats most recent object detection methods in both accuracy and speed. Here we emphasize that our results are directly generated from the off-the-shelf instance mask,
without any box based
training or engineering.

An observation from Figure~\ref{fig:box_speed_ap} is as follows.
If one does not care much about the annotation cost difference between mask
annotation and bounding box annotation, it appears to us that there is no reason to use box detectors for downstream applications, considering the fact that our \method beats most modern detectors in both accuracy and speed.

\subsection{Extension: Panoptic Segmentation}

We also demonstrate the effectiveness of \method on the problem of panoptic segmentation~\cite{KirillovHGRD19}.
The proposed \method can be easily extended to panoptic segmentation by adding the semantic segmentation branch, analogue to the mask feature branch.
We use annotations of COCO 2018 panoptic segmentation task. All models are trained on {\tt train2017}
subset and tested on {\tt  val2017}.
We use the same strategy as in Panoptic-FPN~\cite{kirillov2019panoptic} to combine instance and semantic results.
As shown in Table~\ref{tab:panoptic_segmentation_result}, our method achieves state-of-the-art results and outperforms  other recent box-free methods by a large margin. All methods listed use the same backbone (ResNet50-FPN) except SSAP (ResNet101) and Pano-DeepLab (Xception-71).
Note that UPSNet has used deformable convolutions~\cite{dai2017deformable} for better performance.

\begin{table}[tb]
    \centering
\caption{
\method for \textbf{panoptic segmentation} -- results on MS COCO val2017. $*$ means our re-implementation.
\label{tab:panoptic_segmentation_result}
}
\begin{tabular}{ r c cc}
\toprule
  &  PQ & PQ\textsuperscript{Th} & PQ\textsuperscript{St} \\
\midrule
\emph{box-based:} &&&\\
~~~~AUNet~\cite{li2019attention}       &  39.6 & 49.1 & 25.2  \\
~~~~UPSNet~\cite{xiong2019upsnet}        &  42.5 & 48.5 & 33.4 \\

~~~~Panoptic-FPN~\cite{kirillov2019panoptic} &  39.0 & 45.9 & 28.7   \\
~~~~Panoptic-FPN$^*$-1\X      &  38.7 & 45.9 & 27.8  \\
~~~~Panoptic-FPN$^*$-3\X    & 40.8 & 48.3 & 29.4 \\
\midrule
\emph{box-free:} &&&\\
~~~~AdaptIS~\cite{adaptis}        & 35.9 & 40.3  & 29.3   \\
~~~~SSAP ~\cite{Gao_2019_ICCV}  & 36.5 & $-$ & $-$ \\
~~~~Pano-DeepLab ~\cite{cheng2019panoptic}  & 39.7 & 43.9 & 33.2 \\
~~~~\textbf{\method}             & 42.1 & 49.6 & 30.7  \\
\bottomrule
\end{tabular}
\end{table}

\subsection{Extension: Instance-level Image Matting}

Image matting is a fundamental problem in computer vision and graphics
and has attracted much research attention~\cite{levin2007closed,xu2017deep,lu2019indices}.
Given an image, image matting demands for accurate foreground estimation, which is %
typically
formulated as the alpha map prediction, \ie, to output the soft transition between foreground and background.
Most matting methods
require an extra trimap input, which indicates the regions of absolute foreground, absolute background and unknown.
Few methods~\cite{chen2018semantic,zhang2019late,Liu_2020_CVPR} explore the trimap-free solution.
There is an obvious ambiguity: without trimap it is very challenging
to tell which object is the target foreground object given multiple objects in an image.
However, it can  be solved from another perspective: \textit{to perform image matting at
the instance level.}
Thus, we will be able to deal with arbitrary foreground objects and enable much more flexible, automated image editing.

To further demonstrate the flexibility of the proposed framework, we extend the proposed instance segmentation framework to perform image matting at instance-level.
Thanks to the ability of generating high-quality object masks, our method is able to solve instance-level image matting problem  with minimal modifications.

In this setting, instead of the binary pixel-wise mask, for each instance, we want to attain the soft transitions between objects and the background, \ie, soft matte, which is critically important for photo-realistic image manipulation.

\subsubsection{Method}
\label{sec:soso_method}
The modifications lie in the mask feature branch and the loss function.
In mask feature branch, we use the raw input to enhance the feature representation in image details.
For the loss function, we add a mean average error term computed between the predicted soft matte and the ground-truth matte.
The model is initialized by the weights pre-trained on
COCO instance segmentation.
During inference, we obtain the soft mattes after sigmoid operation and no thresholding is needed.

In parallel with the prediction head, the mask feature branch takes the feature maps as input and fuses them into a \nicefrac{1}{4}
scale feature map.
Specifically, we modify the mask feature branch by introducing the
Details Refinement module.
It takes \nicefrac{1}{4} scale fused features and \nicefrac{1}{1} scale raw image as input and outputs the \nicefrac{1}{1} scale low-level features. Specifically, after 1$\times$1 conv the 4$\times$ bilinear upsampling, the input features are concatenated with the raw RGB input. The output features after three 3$\times$3 convs are expanded to 256 channels through a 1 $\times$ 1 conv. The hidden feature maps all have a small number of channels, \eg, 32.
The output and the upsampled original features are fused together through an element-wise summation as the final mask features, which will be convolved by the predicted convolution kernels to generate individual soft mattes for all the objects.

%

\subsubsection{Dataset}
To our knowledge, no existing public datasets provide separate alpha matte annotations
for each instance.
Thus, we construct a training set consists of human matting data with 880 alpha mattes and 520 images with person category from the LVIS training set~\cite{lvis2019}.

\subsubsection{Results}
We apply the well-trained model to the images collected from the Internet.
As shown in
%
the supplementary, our model is able to generate high-quality
alpha matte.
The predicted alpha matte enables us to perform some image editing applications for instance-level image editing and compositing.
The accurate predictions at object boundaries make the composite image look very natural.
In order to analyze the effect of details refinement, we visualize
the feature maps at its input and output.
As shown in %
the supplementary,
 the input feature maps show
coarse activation at boundaries and details.
 The details are recovered after the detail refinement.

Note that the model trained with human matting data has reasonable matting results for objects of other categories, \eg, the dog in Figure %
\ref{fig:sofi_results} in the supplementary. 
This capability comes from two aspects.
1) As mentioned in Section~\ref{sec:soso_method}, the model is initialized by the weights pre-trained on COCO instance segmentation.
During the training, the weights of the object category branch are frozen for maintaining the strong object recognition ability.
2) The matting capability is transferred from human to other categories. Because in our framework, the matte prediction is class-agnostic.  Basically, the mask branch predicts the soft masks for all the potential objects. The object classes are determined by the category branch.
Thus,
when the model learns with human matting data, the enhanced matte features (see figures in the supplementary)
also benefit objects of other categories.
More results and details of the instance-level image matting are in the supplementary.

%
%
%
%
%

\section{Conclusion}
In this work, we have developed a
simple and
direct instance segmentation framework, termed
SOLO.
To fully exploit the capabilities of this framework, we propose a few variants following the same principle~(\ie, segmenting objects by locations).
Besides the state-of-the-art performance in instance segmentation, we demonstrate the flexibility and effectiveness of SOLO by extending it with minimal modifications to solve object detection, panoptic segmentation and instance-level image matting problems.
Given the simplicity, efficiency, and strong performance of SOLO, we hope that our method can serve as a cornerstone for many instance-level recognition tasks.

\bibliographystyle{ieeetr}
\bibliography{main,supp}

\appendices

\def\Ours{{SOFI}\xspace}
\def\OurMethod{{SOSO}\xspace}

\def\KNN{$k$NN\xspace}

\setcounter{figure}{0}
\setcounter{table}{0}
\renewcommand{\thefigure}{S\arabic{figure}}
\renewcommand{\thetable}{S\arabic{table}}

\vfill

\vspace{4cm}

\section{Instance-level Image Matting}
In this Appendix, we provide more results.

First let us present the new instance-level image matting.

\subsection{Introduction}

Image matting is a fundamental problem in computer vision and graphics
and has attracted much research attention~\cite{levin2007closed,xu2017deep,lu2019indices}.
Given an image, image matting demands for accurate foreground estimation, which is %
typically
formulated as the alpha map prediction, \ie, to output the soft transition between foreground and background.
Most %
matting
methods
require
an extra trimap input, which indicates the regions of absolute foreground, absolute background and unknown. Thus the problem is much simplified into alpha values prediction of the unknown region.
Clearly,
it is a compromise, by putting aside the request of high-level perception and
only focusing on the low-level processing of the %
unknown image
region.
Few methods~\cite{chen2018semantic,zhang2019late} explore the trimap-free solution.
However, they do not generalize well to diverse scenes.
Because there is an obvious ambiguity: without trimap it is
very challenging
to tell which object is the target foreground object given multiple objects in an image.
That is the dilemma of those recent image matting approaches.
However, it %
can
be solved from another perspective: \textit{to perform image matting at
the instance level.}
Thus,
we %
will be
able to deal with arbitrary foreground objects and enable much more flexible, automated image editing.

In parallel, instance segmentation, the problem of recognizing all the objects and localizing them using binary masks, has been widely explored in recent years~\cite{HariharanAGM14,he2017mask,wang2019solo,wang2020solov2}.
Instance segmentation only requires the predicted binary mask to have a relatively high overlap (\eg, IoU = 0.5) with the target.
From
datasets to the evaluation metric, the task of instance segmentation
is highly biased to `detection': to find all the objects with
moderate
localization accuracy.
The developed solutions inevitably fail to %
attain
high-quality segmentation masks for image editing applications.

\begin{figure}[htb]
\centering
\subfigure[Bounding Box]{
\includegraphics[width=0.22\textwidth]{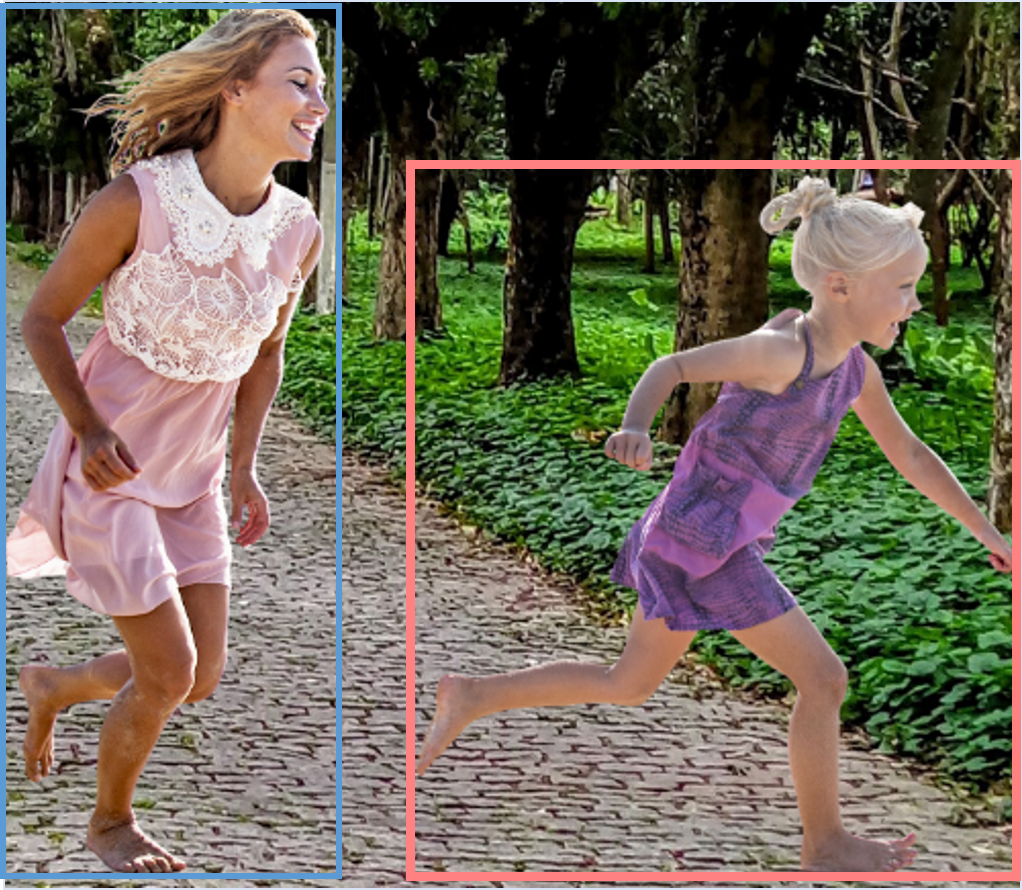}
\label{subfig:3a}}
\hspace{.03in}
\subfigure[Polygon Mask]{
\includegraphics[width=0.22\textwidth]{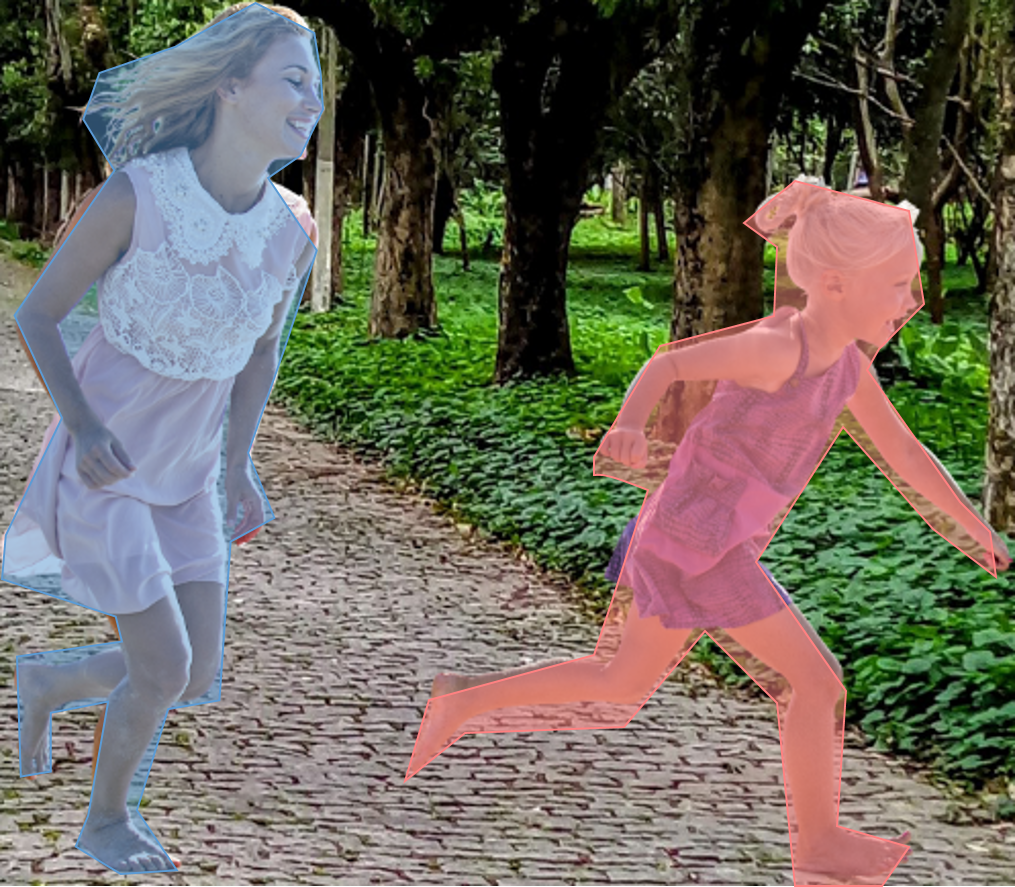}
\label{subfig:3b}}
\hspace{.03in}
\subfigure[Mask]{
\includegraphics[width=0.22\textwidth]{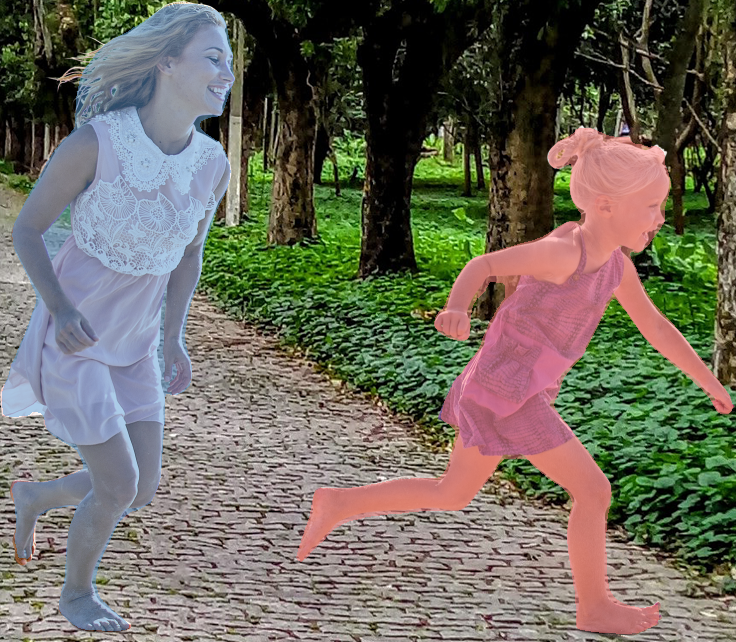}
\label{subfig:3c}}
\hspace{.03in}
\subfigure[Soft Matte]{
\includegraphics[width=0.22\textwidth]{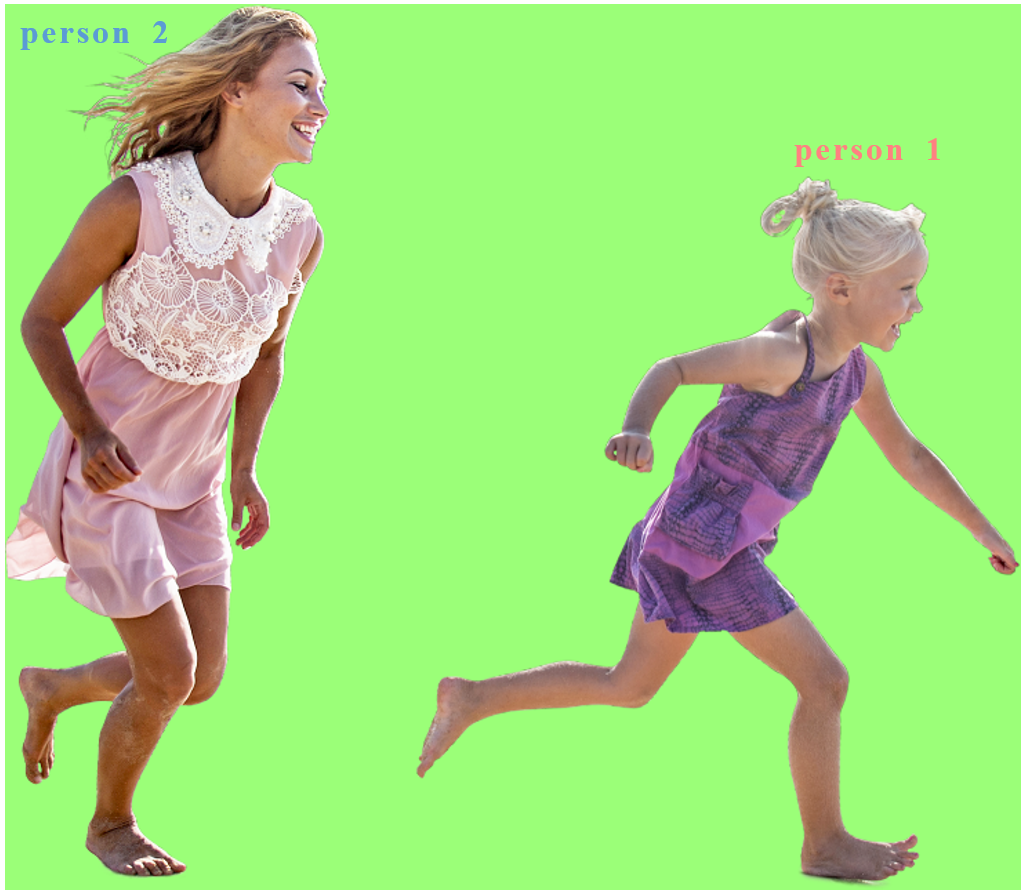}
\label{subfig:3d}}
\caption{Instance-level localization from  coarse representations to fine ones. The problem of object detection evolves into instance segmentation by using binary masks to localize objects instead of coarse bounding boxes, although the masks are usually annotated in the form of polygon masks. We take a step further by proposing \Ours to localize each object with a soft matte, which enables many image manipulation applications.}
\label{fig:fig1}
\end{figure}

In this work, we introduce a new task, termed \textit{soft instance segmentation} (\Ours).
Unlike standard instance segmentation, \Ours requires to output a high-quality soft matte, %
instead of
simply a binary mask.  Unlike traditional image matting,  we require no input of trimaps or manual scratch annotation.
The objective of \Ours is to highly accurately segment all object instances of interest in an image.
Specifically, %
we aim to
attain the soft transitions between objects and the background, which is critically important for photo-realistic image manipulation.
\Ours poses significant challenges as it not only requires high-level perception, \ie discriminating object instances and semantic categories, but also necessitates low-level processing of shape details.
For evaluating the performance clearly and effectively, we carefully design a benchmark tailored for this new task, including datasets and evaluation metrics.

We further propose an effective single-shot architecture customized for the task of \Ours, based on the simple instance segmentation framework SOLO~\cite{wang2020solov2}, termed %
Soft SOLO (\OurMethod).
\OurMethod takes an image as input, directly outputs soft instance mattes and corresponding class probabilities, in a fully convolutional paradigm.
It learns high-level and low-level features and achieves a fusion to perform the final predictions.
We evaluate \OurMethod on both image matting and \Ours.
On image matting, \OurMethod outperforms the state-of-the-art trimap-free approach~\cite{zhang2019late}.
On the proposed \Ours benchmark, we demonstrate that a relative  $30.9\%$ improvement can be achieved over our baseline.

We believe that the proposed task and the accompanying strong method %
open up opportunities to
automated
image manipulation applications, as well as serving as a platform of bridging high-level and low-level visual perception.

\subsection{\Ours Task}

\begin{figure}[t!]
\centering
\subfigure[Image Matting]{
\includegraphics[width=0.43\textwidth]{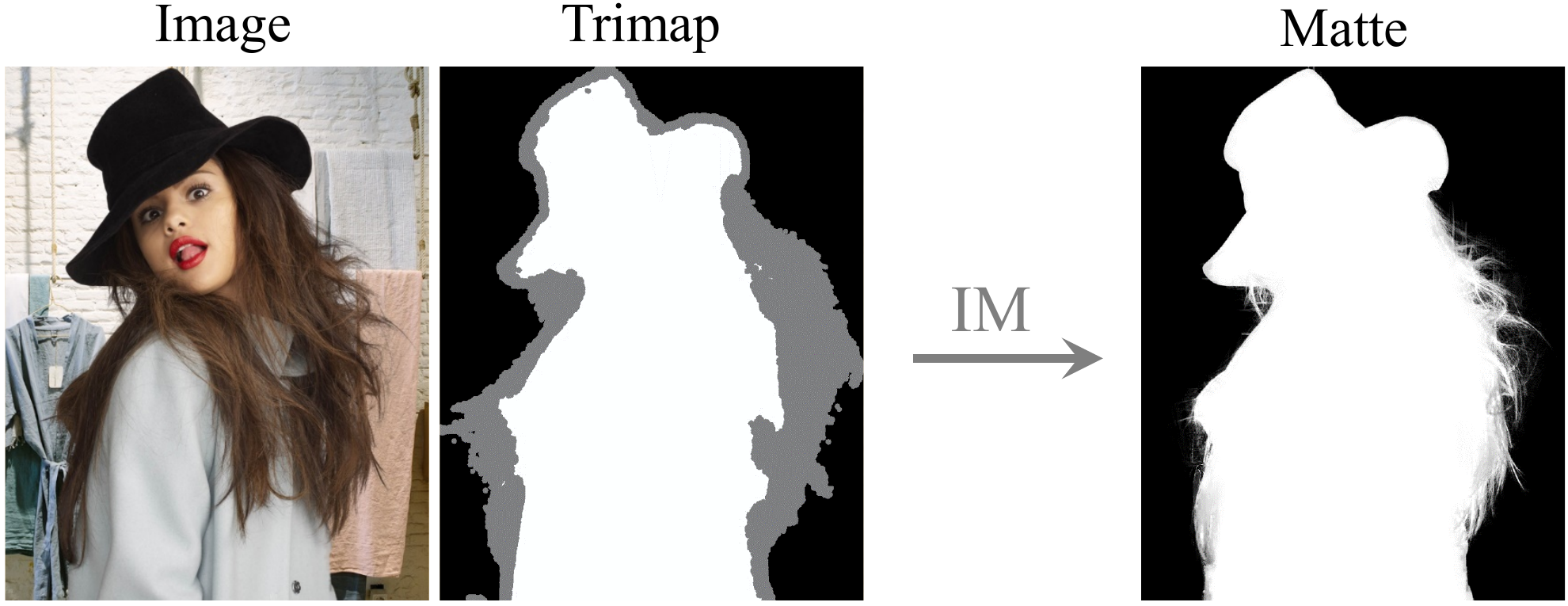}
\label{subfig:pipeline_im}}
\hspace{0.4in}
\subfigure[Soft Instance Segmentation]{
\includegraphics[width=0.43\textwidth]{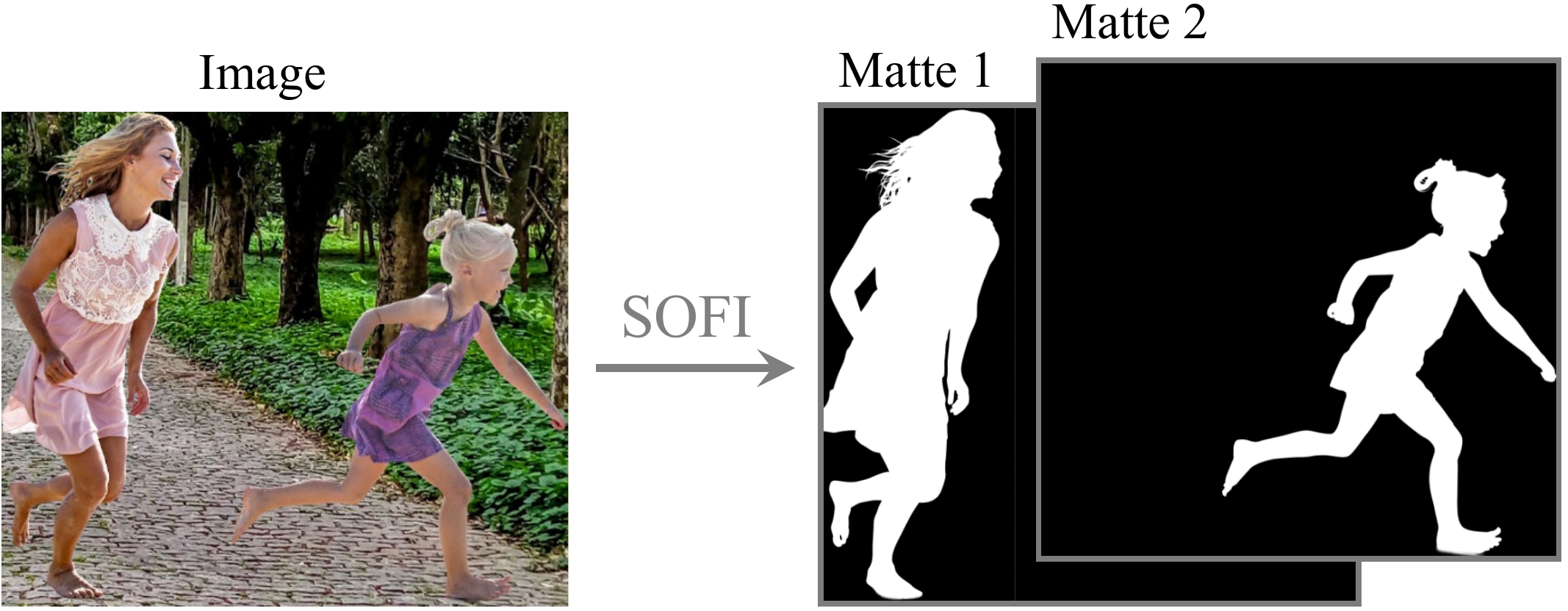}
\label{subfig:pipeline_sofi}}
\caption{Comparison of the pipelines of image matting and our \Ours.}
\label{fig:pipeline_img_sofi}
\end{figure}

\myparagraph{Task Format.}
The format for \Ours is straightforward to define.
Given an input image $ \in \mathbb{ R}^{H\times W\times 3}$ and a predetermined set of $C$ object classes,
the task requires a \textit{\Ours algorithm} to map the input image to a set of soft mattes $\mathbf{M}=\{M_1,\cdots,M_i,\cdots\}$, where $M_i \in \mathbb{ R}^{H\times W\times 1} $, and the corresponding class probabilities $L_i \in \mathbb{ R}^{C}$, where $i$ represents the instance id.
Each pixel value of the soft matte $M_i$ is in the range $[0, 1]$, indicating the transition between the $i^{th}$ object and the background, e.g., 1 for being totally foreground.

\myparagraph{Evaluation Metrics.}
New metrics are proposed for \Ours.
We begin by noting that existing metrics are specialized for either instance segmentation or image matting and can not be used to evaluate the \Ours task.
A suitable metric should be able to simultaneously evaluate high-level per-instance recognition, \ie,  object classification and localization, and low-level per-pixel transition prediction.
    The evaluation metric can be written as follows.
\def\error{{\rm error}}
\begin{align}
\textrm{\Ours-}\error
&=
\frac{1}{|\mathbf{G}|}
\sum_{j=1}^{|\mathbf{G}|} \min_{M_i \in \mathbf{M}} \error (G_j, M_i)
\notag
\\
&+
\frac{1}{|\mathbf{M}|}
\sum_{i=1}^{|\mathbf{M}|} \min_{G_j \in \mathbf{G}} \error (G_j, M_i),
\label{eq:metric}
\end{align}
where  $G_j$ represents the soft matte of $j^{th}$ object of the %
in total
$|\mathbf{G}|$ ground-truth objects; $M_i$ represents the $i^{th}$ predicted soft matte. Here $\error(\cdot, \cdot )$ represents an error function, \eg, mean squared error.
The calculated errors are averaged over the $C$ object classes and all the test images.
We adopt the widely used error functions Mean Squared Error (MSE) and Sum of Absolute Differences (SAD) in image matting as the $\error$ to instantiate the \Ours-$\error $, %
which are
\Ours-MSE and \Ours-SAD.

\myparagraph{Dataset.}
To our knowledge no existing public dataset  provides  separate alpha matte annotations for each instance. To evaluate the performance of SOFI methods, we construct a \Ours dataset using a human matting dataset (details in supplementary) and a gallery of collected background images.
Specifically, multiple foreground persons are composited with a selected background image by carefully designed rules, and the corresponding alpha mattes are collected and aligned, as the instance-level alpha matte annotations.
We manually filter out bad cases like objects with too small visible part and severe occlusion.
The dataset contains %
200 images and 537 high-quality alpha mattes in total, termed
\Ours-200.
As for the training set, it consists of two parts, human matting data with 880 alpha mattes and 520 images with person category from the LVIS~\cite{lvis2019} training set.
More details about the datasets can be referred to the supplementary.

\subsection{Our Method: Soft SOLO}
We present the details of the proposed Soft SOLO design
in this section.

\begin{figure*}[t]
  \centering
\includegraphics[width=0.7048\textwidth]{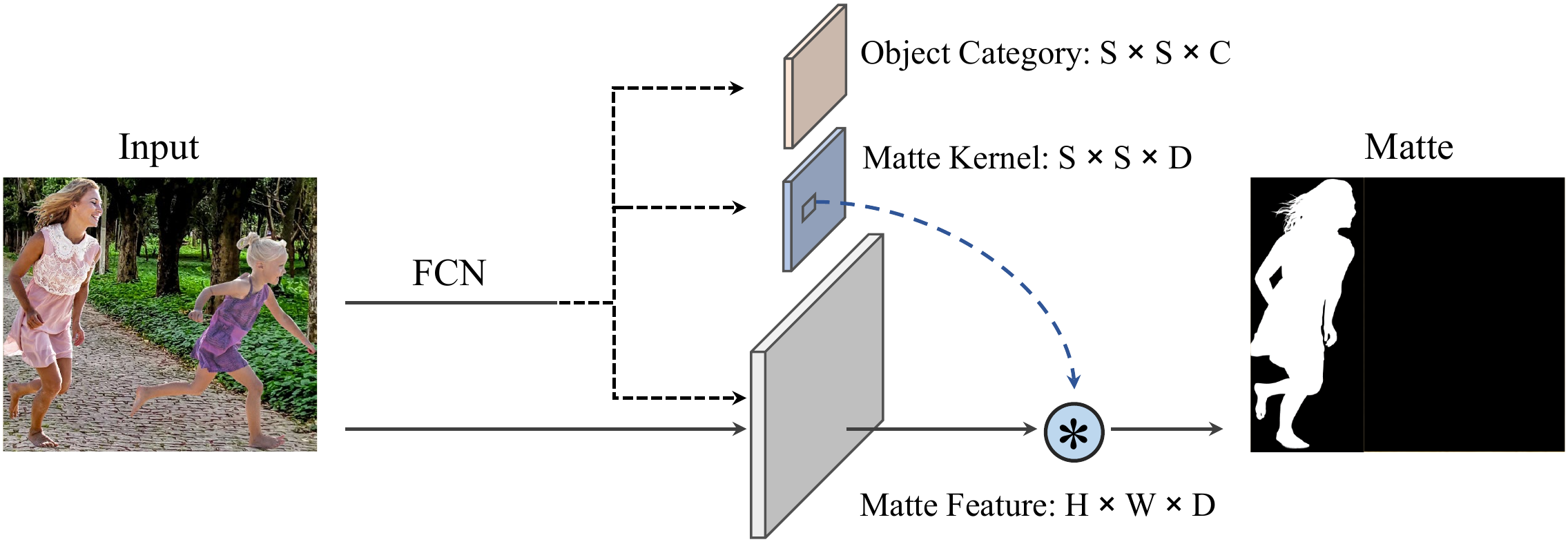}
\vspace{0.1in}
  \caption{\textbf{Overview of \OurMethod.} \OurMethod directly maps the input image to desired object categories and the corresponding soft mattes. `$\circledast$' denotes convolution.}
 \label{fig:soso_pipeline}
\end{figure*}

\subsubsection{Network Architecture}
Our architecture is built upon
SOLO, specifically, the SOLOv2 framework.
We show an overview of \OurMethod pipeline in Figure~\ref{fig:soso_pipeline}.
After a convolutional backbone network,
FPN~\cite{lin2017feature} is adopted to generates a pyramid of feature maps with different sizes with  a  fixed number of channels for each level.
Then prediction head including object category branch and matte kernel branch is attached on the pyramid of of feature maps.
Weights for the head are shared across different levels.
In parallel with the prediction head, matte feature branch takes the feature maps as input and fuses them into a 1/4 scale feature map.
We redesign the matte feature branch by introducing the Details Refinement module (DR).
Details Refinement module takes 1/4 scale fused features and 1/1 scale raw image as input and outputs the 1/1 scale low-level features.
Specifically, after 1$\times$1 conv the 4$\times$ bilinear upsampling, the input features are concatenated with the raw RGB input.
The output features after three 3$\times$3 convs are expanded to 256 channels through a $1 \times 1$ conv.
The feature maps in Details Refinement module all have a small number of channels, \eg, 32.
The design is motivated by the observation of the mask feature behavior in SOLOv2: the feature maps are either position-aware or responsible for activating all the foreground objects.
It enables us to simply extract the small part of global activation feature maps to perform low-level details learning.
The output of Details Refinement module and the upsampled original features are fused together through an element-wise summation as the final matte features, which will be convolved by the predicted convolution kernels to generate individual soft mattes for all the objects.

\subsubsection{\OurMethod Learning}
The label assignment is the same as SOLO, except that the ground truths of mask branch are no longer binary masks, but soft mattes.
The training loss function is made up of two components, defined as:
\begin{equation}
\label{eq:loss_all}
L = L_{cate} + L_{matte},
\end{equation}
where $L_{cate}$ is the focal loss~\cite{LinGGHD17} for object category classification.
$L_{matte}$ is the loss for soft matte prediction:
\begin{equation}
\label{eq:loss_matte}
L_{matte} = L_{mae} + L_{dice},
\end{equation}
where $L_{mae}$ is the mean average error and $L_{dice}$ is the Dice loss.
Losses for soft matte prediction are normalized by the number of instances.
The model is initialized by the weights pre-trained on MS COCO~\cite{lin2014microsoft} instance segmentation.
During the training, the weights of object category branch is frozen for maintaining the strong object recognition ability.
We adopt a mixed training strategy, which enables us to train a strong model using off-the-shelf datasets.
Both the human matting data and well-annotated instance segmentation data are used.
The former provides high-quality matte supervision, while the latter helps maintain the ability to distinguish instances.

\subsubsection{Inference}
During inference, the prediction branches execute in parallel after the backbone network and FPN, producing category scores, predicted convolution kernels and matte features.
For category score
$\mathbf{L}_{i,j}$ at grid $(i, j)$.
a confidence threshold of $0.1$ is used to filter out predictions with low confidence.
For each valid grid, the corresponding predicted matte kernels are used to perform convolution on the matte features.
We %
obtain
the soft mattes after \texttt{sigmoid} operation.
Finally, we employ the Matrix NMS~\cite{wang2020solov2} to remove redundant predictions.

\subsection{Experiments}
We present quantitative and qualitative results on the proposed \Ours-200 benchmark.
Besides, we apply the model developed on our \Ours task to instance segmentation and image matting respectively, and show clear comparisons with their state-of-the-art approaches.
A detailed ablative analysis shows how each component contributes to the results.

\myparagraph{Experiment settings.}
The models in this work are trained with stochastic gradient descent (SGD).
We use synchronized SGD over 8 GPUs with a total of 8 images per mini-batch (1 image per GPU).
All models are trained for 6 epochs with an initial learning rate of 0.0005, which is then divided by 10 at 5th epoch.
Weight decay of 0.0001 and momentum of 0.9 are used.
All models are initialized from COCO pre-trained weights.
The shorter image side is fixed to 800 pixels.
Unless otherwise specified, we use ResNet-101~\cite{he2016deep} as backbone for the compared models and ours.

\subsubsection{Main Results}
The results on \Ours are reported in Table~\ref{tab:main_results}.
We compare two baseline methods, SOLOv2 and the proposed \OurMethod, in the same training settings.
We also show that how Mask R-CNN and PointRend perform in the challenging \Ours task, for presenting a broader comparison.
Both trained on COCO dataset for instance segmentation, SOLOv2 performs much better than Mask R-CNN and PointRend.
This is easy to understand: Mask-RCNN’s mask head is typically restricted to $28 \times 28$  resolution, resulting in coarse mask predictions.
However, this defect is ignored in instance segmentation, as the task only requires the predicted binary mask to have a relatively high overlap with the target.
It shows that how necessary it is to introduce a new task and the corresponding benchmark, for more detailed and accurate instance segmentation.

Training SOLOv2 on the matting dataset brings slight improvements.
Using the same training setting, \OurMethod achieves a significant boost, which is a $30.9\%$ relative improvement.
It demonstrates that our solution could serve as a solid and effective baseline for this new and challenging \Ours task.

\begin{table}[t]
    \centering
     \caption{\textbf{\Ours results}  on the \Ours-200 dataset. SOLOv2$^*$ is the model trained using the same schedule and training data as \OurMethod.}
    \footnotesize
    \begin{tabular}{l |cc}
         &\Ours-MSE ($\times 10^{-3}$) & \Ours-SAD ($\times 10^{3}$)  \\
        \Xhline{1pt}
        ~~Mask R-CNN~\cite{he2017mask} & 24.70 & 22.99 \\
        ~~PointRend~\cite{kirillov2019pointrend} & 19.65 & 18.57 \\
        ~~SOLOv2~\cite{wang2020solov2}     & 12.13 & 12.06   \\
        ~~SOLOv2$^*$  & 10.21 & 10.33  \\
        ~~\textbf{\OurMethod}   & \textbf{7.06} & \textbf{9.24}   \\
    \end{tabular}
     \label{tab:main_results}
\end{table}

\begin{figure*}[b!]
    \centering
    \subfigure[Image Matting Results]{
        \includegraphics[width=0.72646\textwidth]{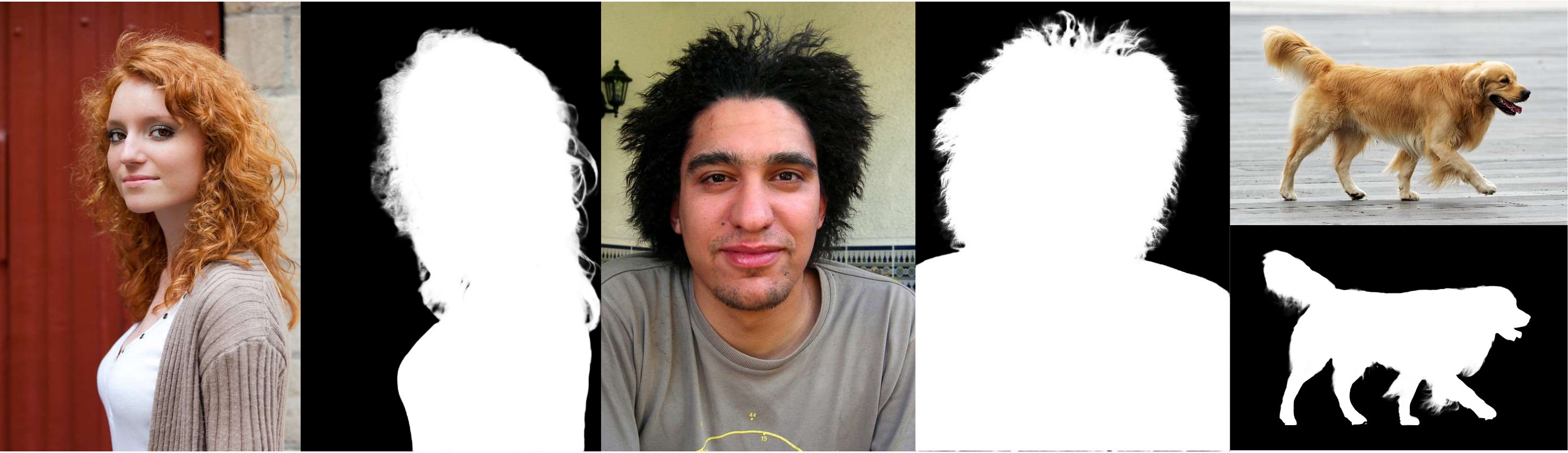}
        \label{fig:single_matte}
    }
    \subfigure[Image Editing Applications]{
        \includegraphics[width=0.72646\textwidth]{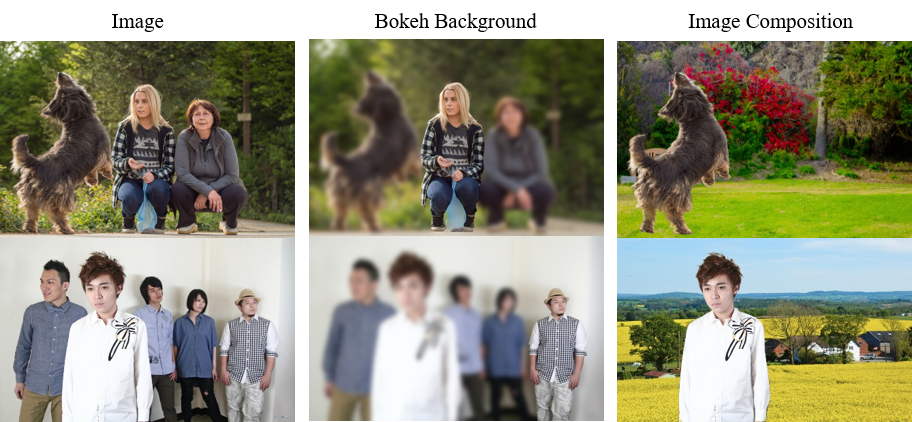}
        \label{fig:img_editing}
    }
\caption{(a) Visualization of image matting results. We show the input image and the corresponding output
alpha matte of our method. Best viewed on screens. (b) Demonstration of image editing applications (from left to right: original image; Bokeh background
effect focusing on one instance; image composition).}
\label{fig:sofi_results}
\end{figure*}

\subsubsection{Comparisons with Image Matting Methods}
The solution developed on the proposed \Ours task can be directly applied to the classical image matting problem, which can be viewed as a special case of \Ours, \ie, \Ours with a single object.
In Table~\ref{tab:matting_results}, we report image matting results on human matting dataset and compare with a few trimap-based image matting methods: \KNN matting~\cite{chen2013knn}, closed-form matting~\cite{levin2007closed}, large kernel matting~\cite{he2010fast} and random walk matting~\cite{grady2005random}.
Note that \textit{those methods use ground-truth trimaps as input and only need to deal with the unknown region}. Therefore, we compare the results over the unknown regions.
\textit{Our method yields competitive results on all the four metrics, using only the raw RGB image as input.}

Furthermore, we compare with the recent trimap-based image matting method LFM~\cite{zhang2019late}.
LFM~\cite{zhang2019late} takes an RGB image as input and predicts a single soft matte without instance labels. We calculate the evaluation metrics over the whole image.
As shown in Table~\ref{tab:matting_results_trimapfree}, our method shows large superiority in three out of four metrics.

\begin{table}[h]
    \centering
     \caption{\textbf{Image matting results}  on the human matting dataset. T means that the ground-truth trimap is %
     provided to the algorithm
     as input. The metrics are calculated over the unknown regions.
     Ours is among the best ones with no trimaps needed.
     }
     \footnotesize
    \scalebox{0.9}{
    \begin{tabular}{l |cccc}
         & MSE ($\times 10^{-3}$) & SAD ($\times 10^{3}$) & Grad.\  ($\times 10^{1}$) & Conn.\  ($\times 10^{3}$) \\
        \Xhline{1pt}
        ~~T +~\cite{levin2007closed}    & 20.16 & 4.87  & 7.34 & 4.43  \\
        ~~T +~\cite{chen2013knn}     &  25.67 & 5.66 & 8.51 & 5.29  \\
        ~~T +~\cite{he2010fast}     & 41.04  & 8.56  & 11.08 & 8.19  \\
        ~~T +~\cite{grady2005random}     & 48.07  & 6.45  & 15.34 & 6.24  \\
        \Xhline{1pt}
        ~~\textbf{\OurMethod}      & 28.51 & 4.73 & 9.14 & 4.61  \\
    \end{tabular}}
     \label{tab:matting_results}
\end{table}

\begin{table}[h]
    \centering
     \caption{\textbf{Trimap-free image matting results}  on the human matting dataset. The metrics are calculated over the entire image. }
     \footnotesize
    \scalebox{0.88}{
    \begin{tabular}{l |cccc}
         & MSE ($\times 10^{-3}$) & SAD ($\times 10^{3}$) & Grad.\  ($\times 10^{1}$) & Conn.\  ($\times 10^{3}$) \\
        \Xhline{1pt}
        ~~LFM~\cite{zhang2019late} & 14.47 & 13.51  & \textbf{8.66} & 13.20  \\
        ~~\textbf{\OurMethod}      & \textbf{7.18} & \textbf{7.28}  & 11.21  & \textbf{7.05}   \\
    \end{tabular}}
     \label{tab:matting_results_trimapfree}
\end{table}

\begin{figure*}[hbt]
\centering
\subfigure[Before the Details Refinement module]{
\includegraphics[width=0.4328\textwidth]{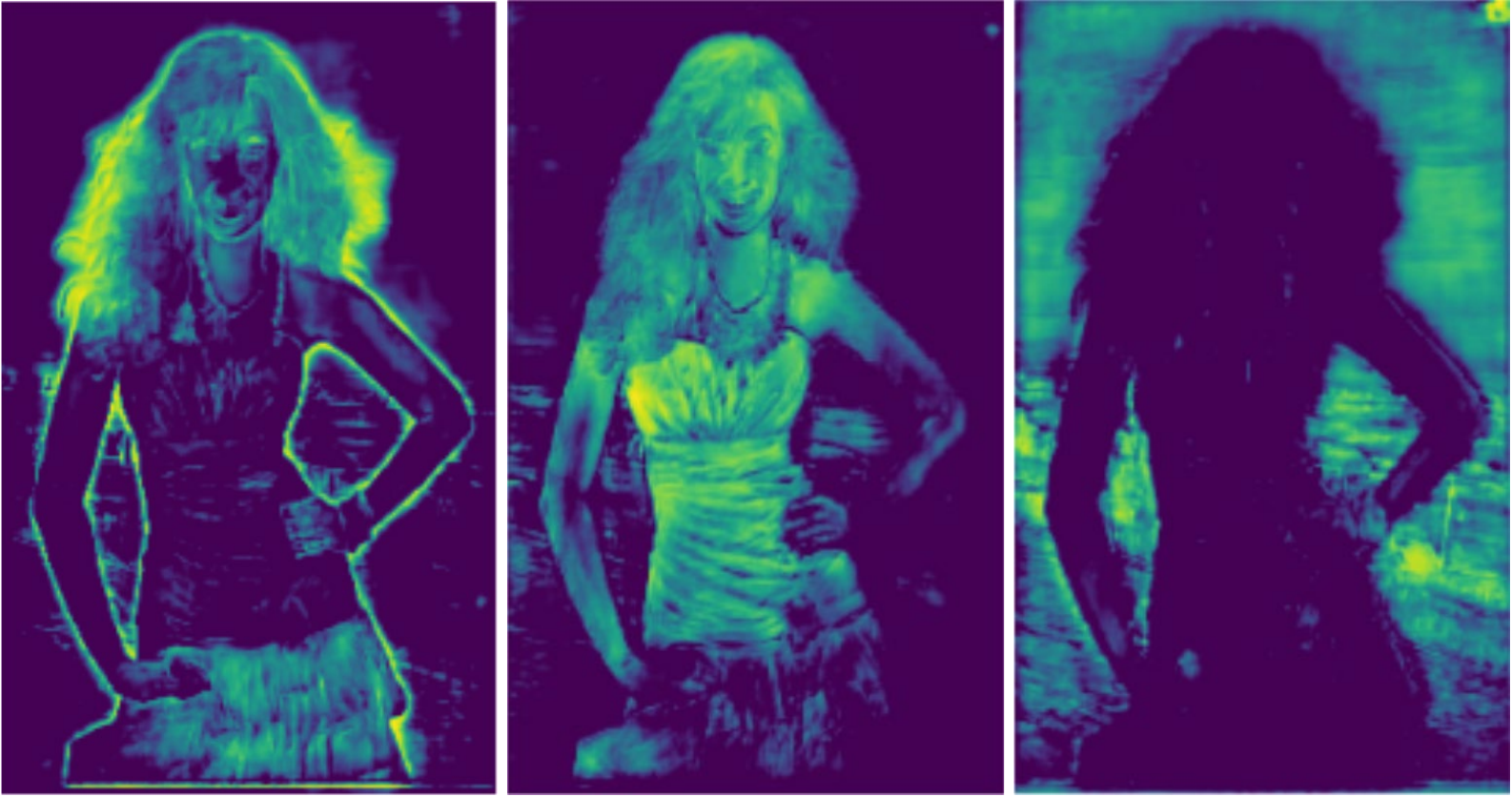}
\label{fig:llm_input}
}
\hspace{.03in}
\subfigure[After the Details Refinement module]{
\includegraphics[width=0.4328\textwidth]{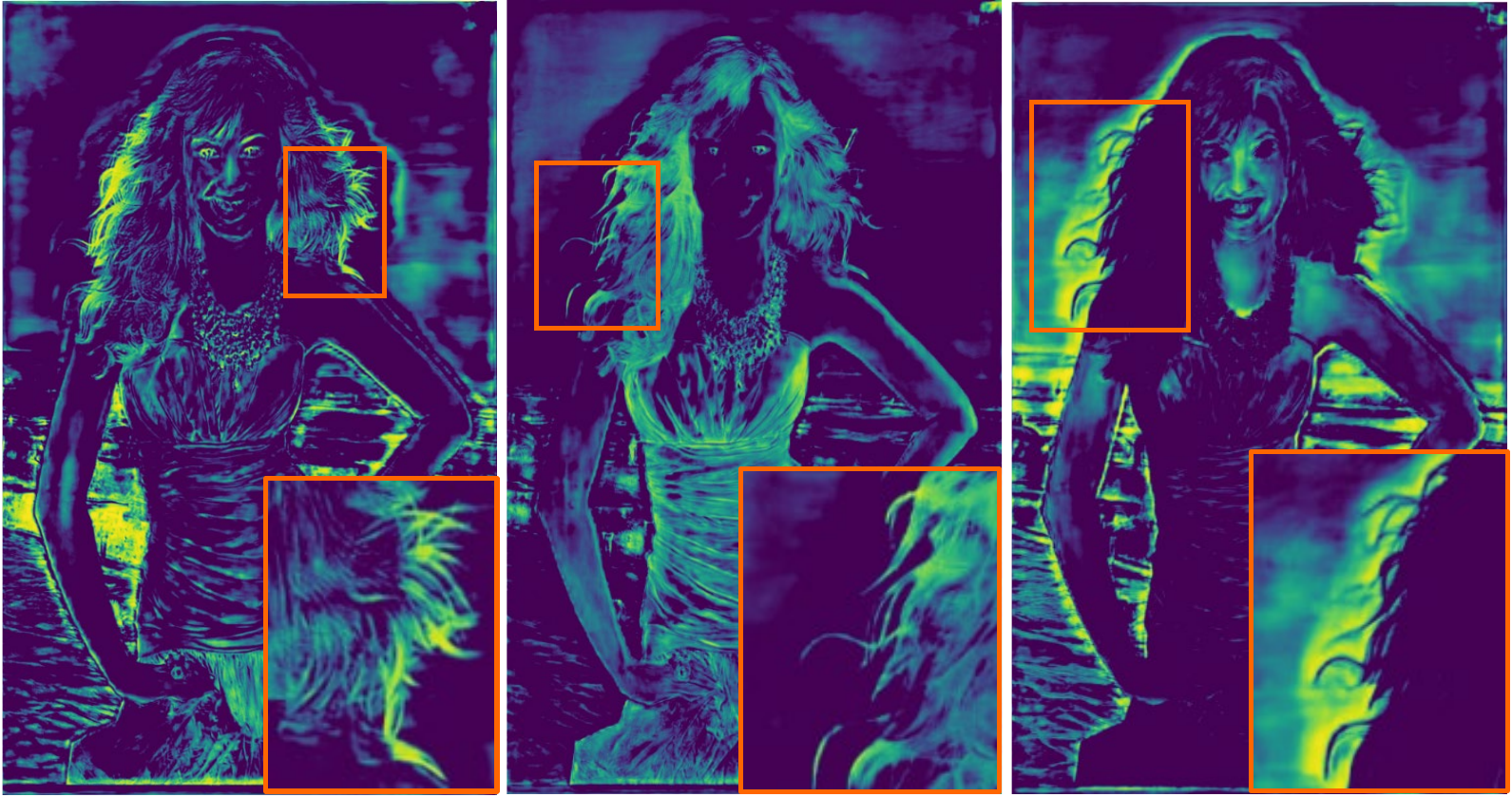}
\label{fig:llm_output}
}
\caption{Behavior of the Details Refinement module. Each plotted sub-figure corresponds to one of the 32 channels of the input/output feature maps.
The output feature map recovers the low-level details. Best viewed on screens.}
\label{fig:vis_llm}
\end{figure*}

\subsubsection{Ablative Analysis}
We present ablative experiments on the human matting validation set.
Table~\ref{table:sofi_ablation}(a) shows the
 quantitative
effect of the Details Refinement module.
The results drop sharply when there is no raw RGB input to the Details Refinement module,
demonstrating %
that
the low-level cues from the raw input play a significant role for %
extracting
low-level details.
In Table~\ref{table:sofi_ablation}(b), we show the effect of different loss components by training several models turning them on and off.
When only single loss term is used,  MAE loss shows advantages over Dice loss.
The model with both loss functions achieves the best performance.
An interesting fact is that Dice loss itself can also achieve good results on those matting metrics.
Although it is proposed and used for per-pixel binary classification for years,
we show that it could be directly applied to the task of alpha matte prediction and achieve promising results.

\begin{table}[hbt]
\caption{Ablation experiments on the human matting validation set.}
\vskip 0.05 in
\begin{minipage}{0.9\linewidth}
\begingroup
\centering
\footnotesize{ (a) \textbf{Details Refinement module.} The RGB input plays a key role in the Details Refinement module.}
\endgroup
\vskip 0.05 in
\centering
\scalebox{0.9}{
\begin{tabular}{l|cccc}
         & MSE  & SAD & Grad  & Conn \\
        \Xhline{1pt}
        \OurMethod &  37.44 & 5.18 & 10.99 & 5.02  \\
        ~$-$ w/o RGB input &  39.11  & 5.39 & 11.36 & 5.27 \\
        ~$-$ w/o feature fusion &  39.88 & 5.50  & 11.85 & 5.38  \\
    \end{tabular}}
\end{minipage}
\vskip 0.05 in
\begin{minipage}{.9\linewidth}
\begingroup
\footnotesize{ (b)\textbf{ Loss components.} We use both loss terms for the better results.}
\endgroup
\vskip 0.05 in
\centering
\scalebox{0.9}{
\begin{tabular}{p{.9cm}<{\centering} p{1cm}<{\centering} |cccc}
        $L_{mae}$ & $L_{dice}$ & MSE  & SAD & Grad & Conn \\
        \Xhline{1pt}
         \cmark & &   38.31  & 5.15  & 11.26 & 5.02 \\
        & \cmark &  39.11 & 6.35  & 12.19 & 6.03 \\
        \cmark & \cmark & 37.44  & 5.18 & 10.99 & 5.02  \\
    \end{tabular}
}
\end{minipage}
\label{table:sofi_ablation}
\end{table}

\section{Results on Object Detection}

We show results in Table \ref{table:box_det}
on using SOLO for bounding-box based object detection.

\begin{table*}[t!]
\centering
\small
\caption{\textbf{Object detection} box AP (\%) on
the
MS COCO \texttt{test}-\texttt{dev}. Although our bounding boxes are directly generated from the predicted masks, the accuracy outperforms most state-of-the-art methods.
}
\footnotesize
\begin{tabular}{ r  |l|cccccc}
  & backbone  & AP & AP$_{50}$ & AP$_{75}$ & AP$_S$ & AP$_M$ & AP$_L$  \\
\Xhline{1pt}
YOLOv3~\cite{yolov3}       & DarkNet53   & 33.0 & 57.9 & 34.4 & 18.3 & 35.4 & 41.9 \\
SSD513~\cite{ssd}          & ResNet-101  & 31.2 & 50.4 & 33.3 & 10.2 & 34.5 & 49.8 \\
DSSD513~\cite{ssd}         & ResNet-101  & 33.2 & 53.3 & 35.2 & 13.0 & 35.4 & 51.1 \\
RefineDet~\cite{refinedet} & ResNet-101  & 36.4 & 57.5 & 39.5 & 16.6 & 39.9 & 51.4 \\
Faster R-CNN~\cite{fpn}    & Res-101-FPN & 36.2 & 59.1 & 39.0 & 18.2 & 39.0 & 48.2 \\
RetinaNet~\cite{focalloss} & Res-101-FPN & 39.1 & 59.1 & 42.3 & 21.8 & 42.7 & 50.2 \\
FoveaBox~\cite{foveabox}      & Res-101-FPN & 40.6 & 60.1 & 43.5 & 23.3 & 45.2 & 54.5 \\
RPDet~\cite{reppoints}         & Res-101-FPN & 41.0 & 62.9 & 44.3 & 23.6 & 44.1 & 51.7 \\
FCOS~\cite{fcos}           & Res-101-FPN & 41.5 & 60.7 & 45.0 & 24.4 & 44.8 & 51.6 \\
CenterNet~\cite{centernet} &Hourglass-104& 42.1 & 61.1 & 45.9 & 24.1 & 45.5 & 52.8 \\
\hline
\textbf{\method}  & Res-50-FPN       & 40.4 & 59.8 & 42.8 & 20.5 & 44.2 & 53.9 \\
\textbf{\method}  & Res-101-FPN      & 42.6 & 61.2 & 45.6 & 22.3 & 46.7 & 56.3 \\
\textbf{\method}  & Res-DCN-101-FPN  & 44.9 & 63.8 & 48.2 & 23.1 & 48.9 & 61.2 \\
\end{tabular}
\label{table:box_det}
\end{table*}

\def\visimgheighta{3.08cm}

\def\visimgheightf{2.955cm}

\def\visimgheightg{3.11cm}

\begin{figure*}[t!]
\centering 
{
\includegraphics[height=\visimgheighta]{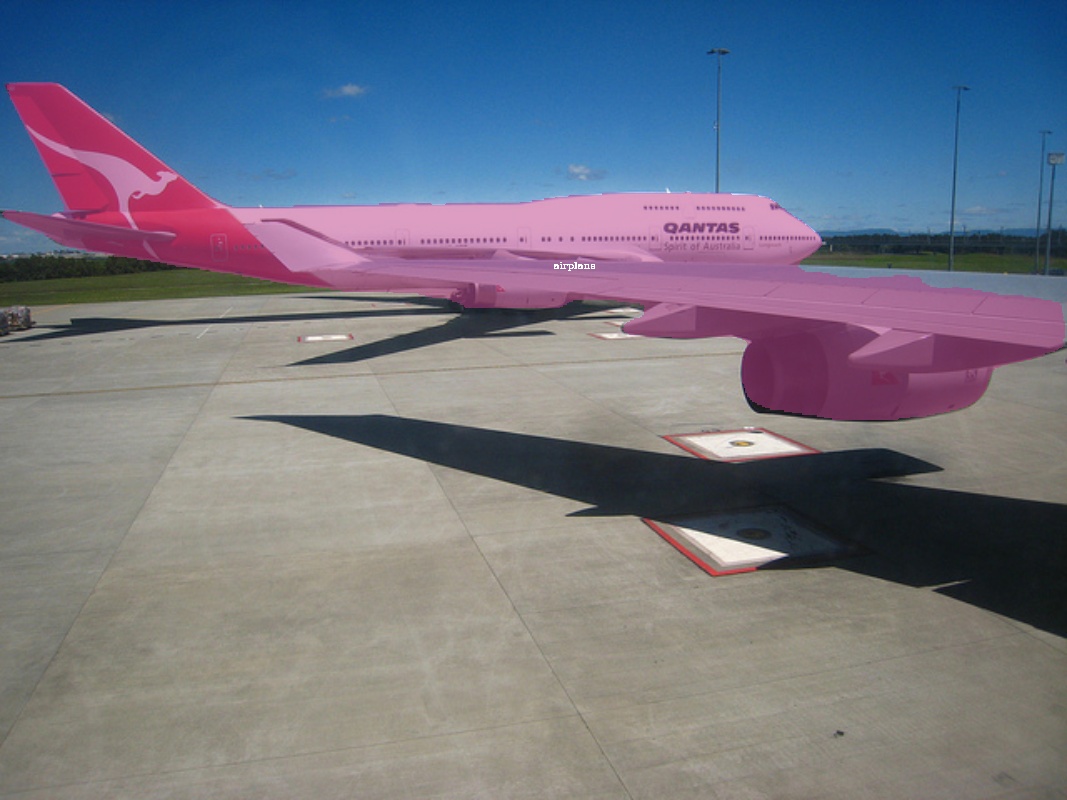}
\includegraphics[height=\visimgheighta]{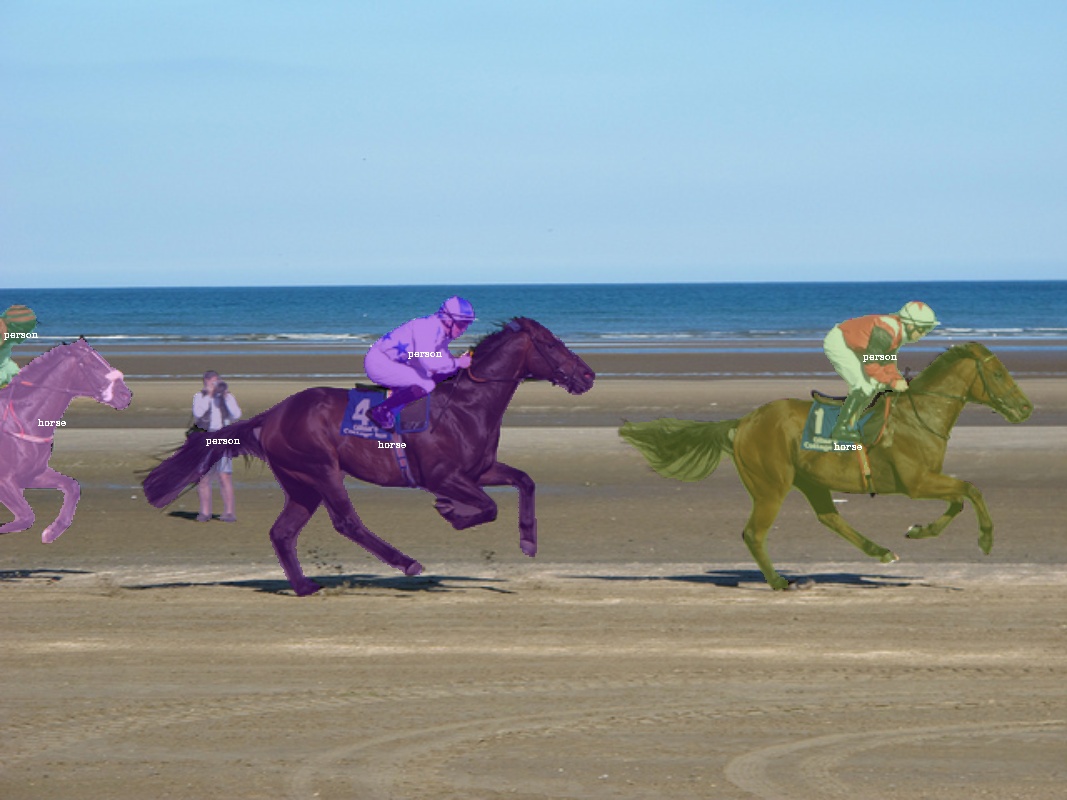}
\includegraphics[height=\visimgheighta]{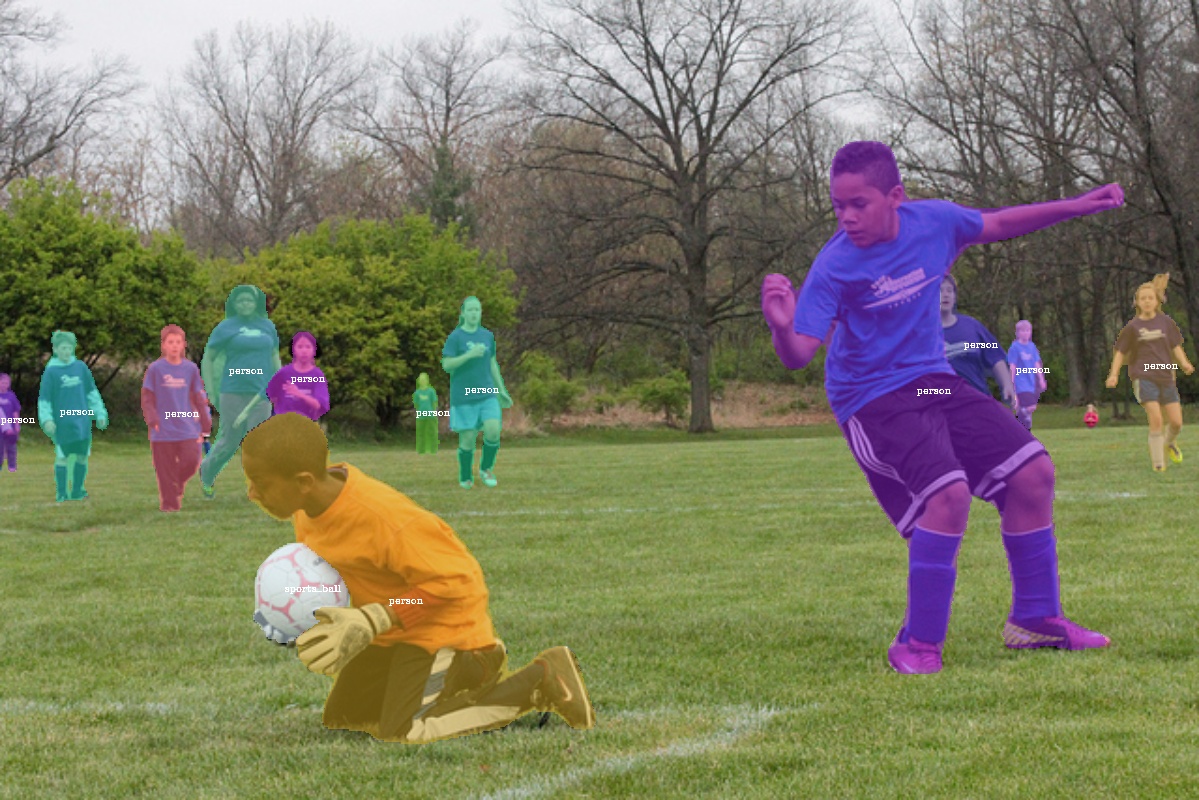}
\includegraphics[height=\visimgheighta]{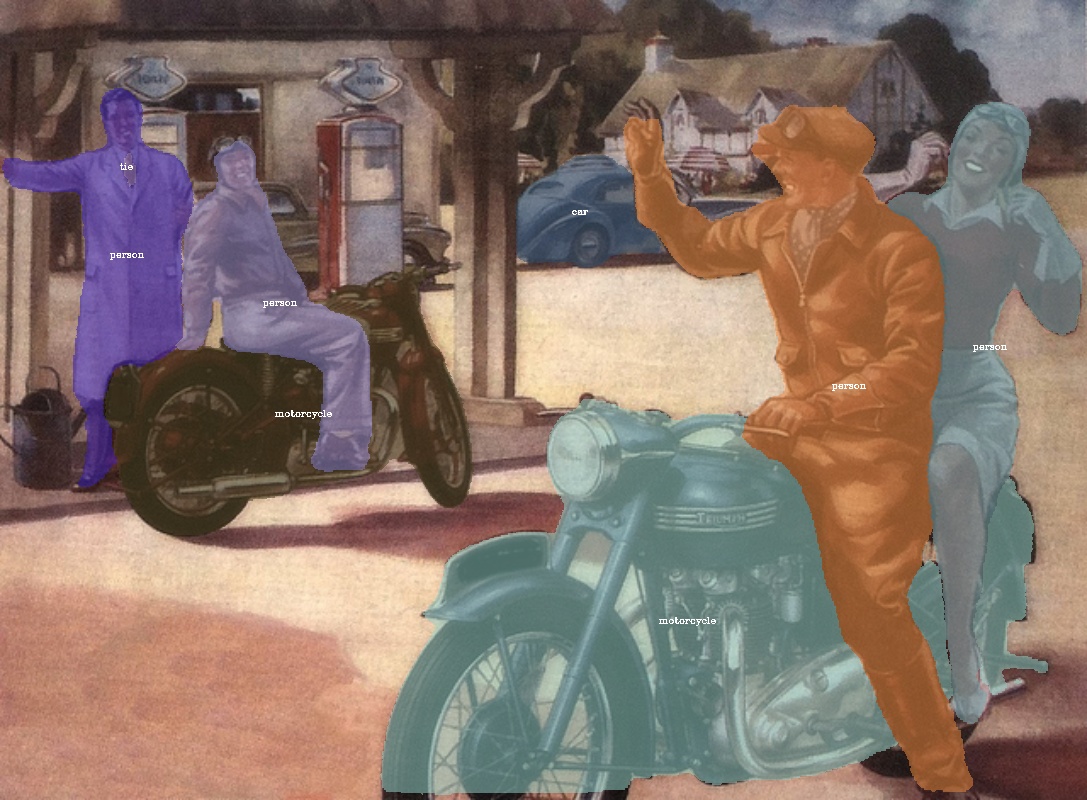}

\includegraphics[height=\visimgheightf]{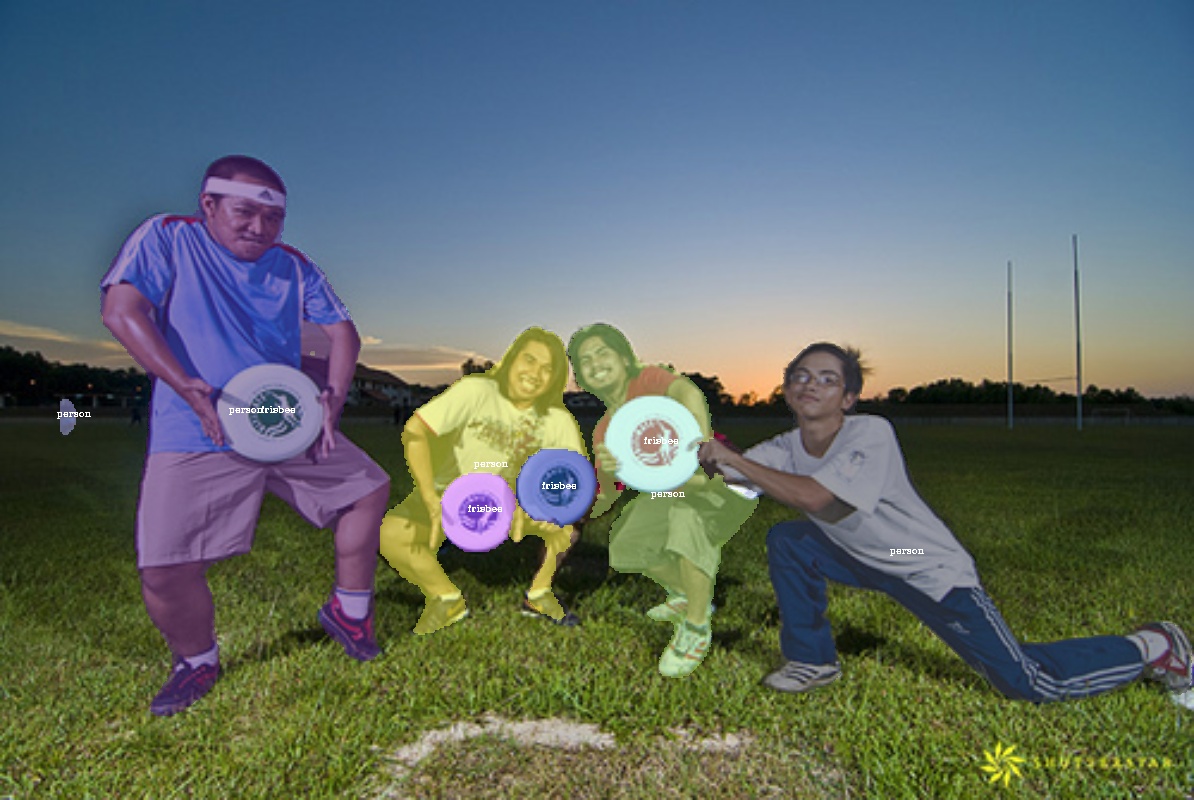}
\includegraphics[height=\visimgheightf]{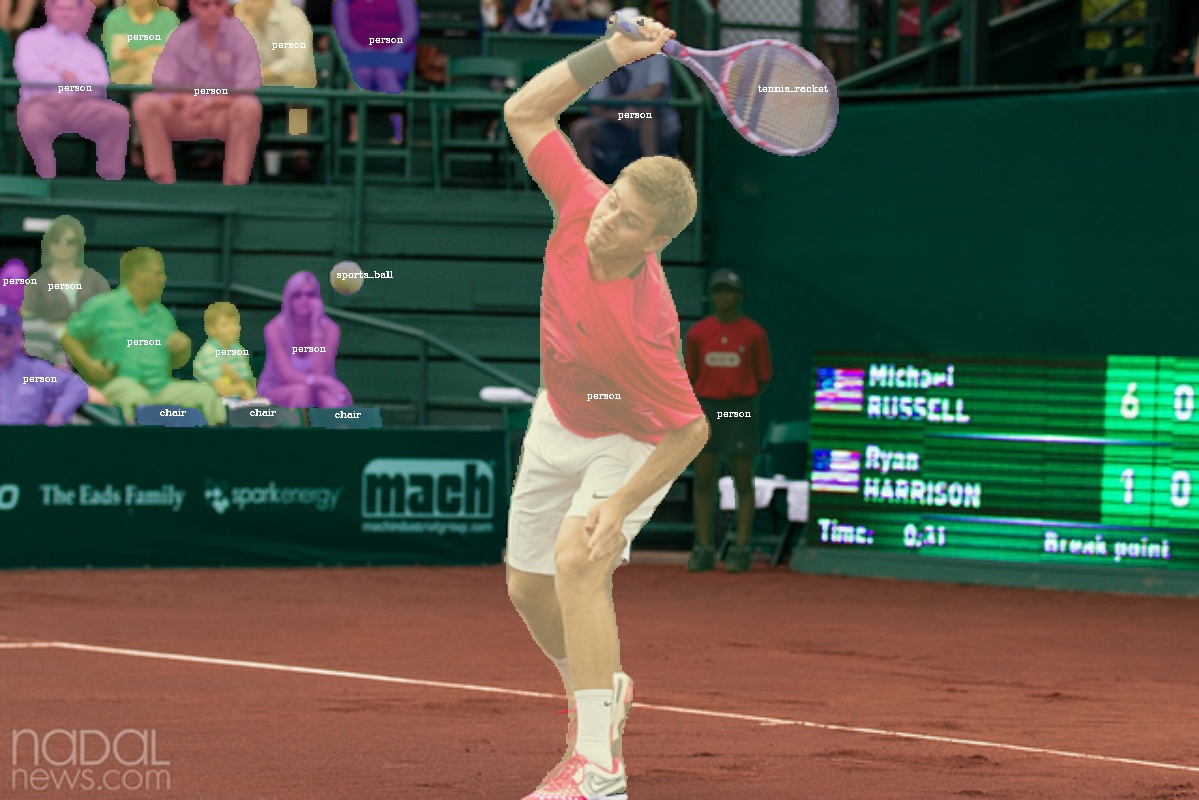}
\includegraphics[height=\visimgheightf]{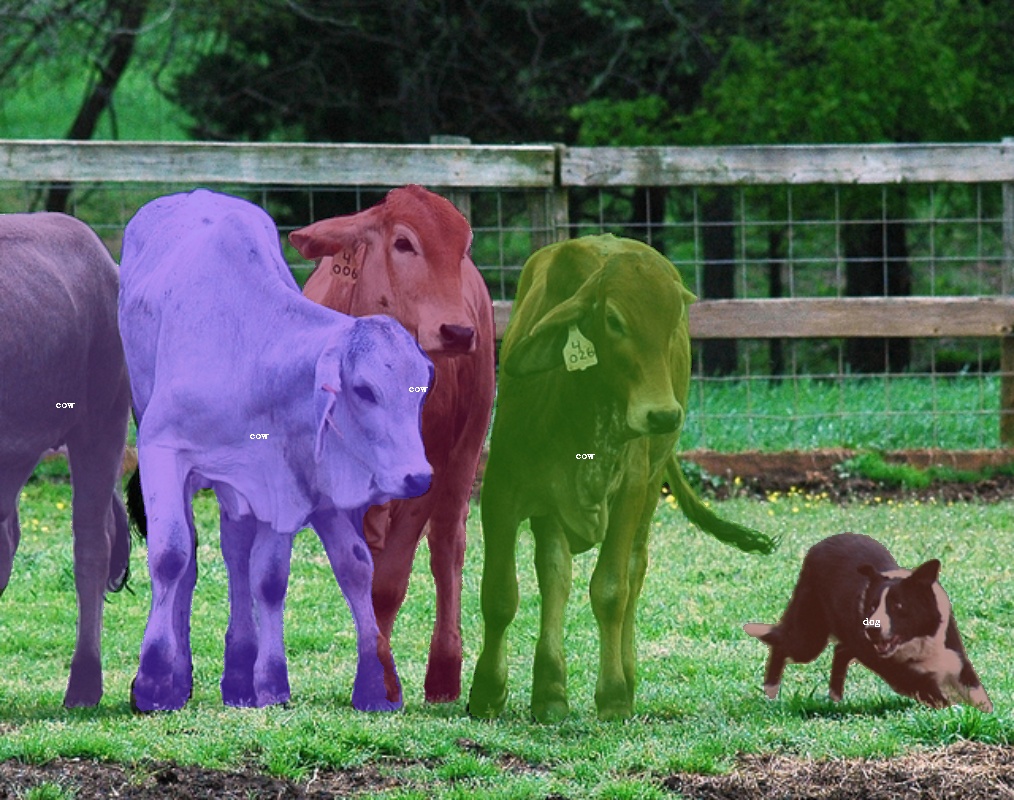}
\includegraphics[height=\visimgheightf]{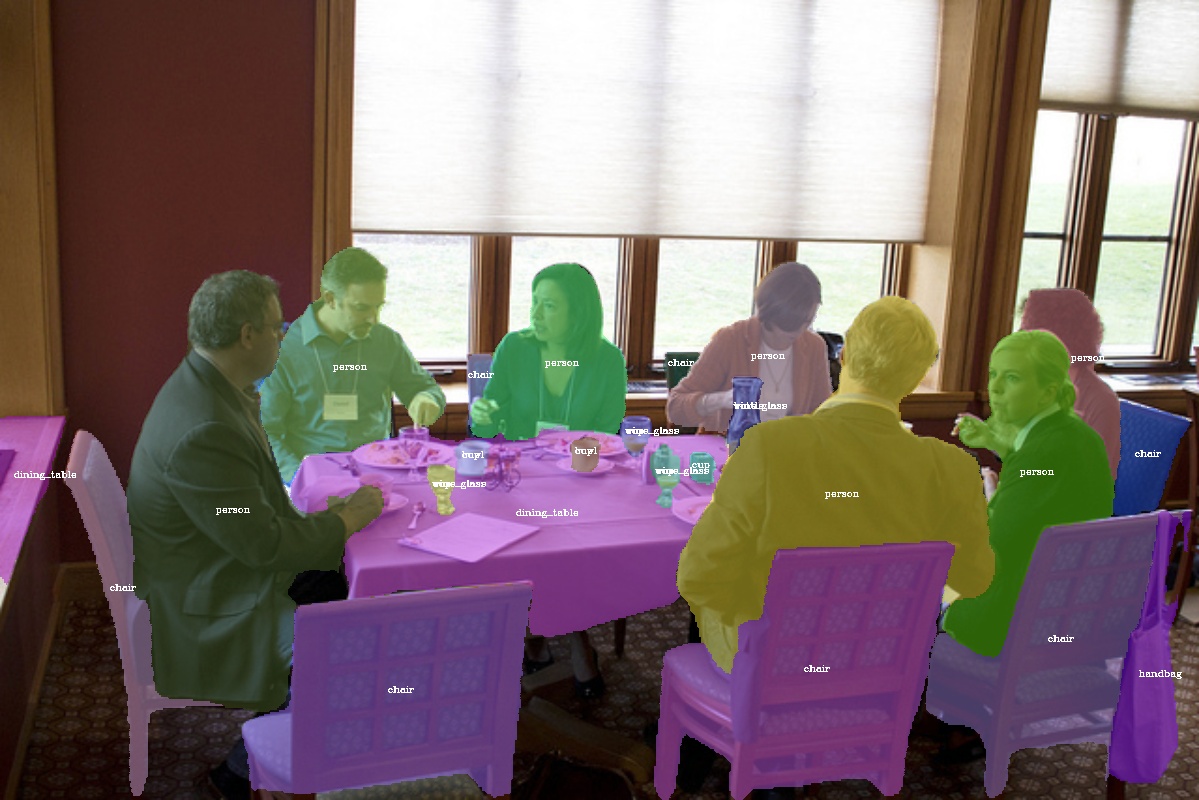}

} %
\caption{\textbf{Visualization of instance segmentation results}
using SOLOv2 ResNet-101 backbone.
The model is trained on the COCO \texttt{train2017} dataset, achieving a mask AP of 39.7\%  on the COCO \texttt{test}-\texttt{dev}.
}
\label{fig:Vis}
\end{figure*}

\section{Visualization}
We show visualization of the final instance segmentation results in Figure~\ref{fig:Vis}. Different objects are in different colors. Our method shows  promising results in diverse scenes. It is worth pointing out that the details at the boundaries are segmented well, especially for large objects.

\begin{figure*}[hbt]
\centering
\subfigure[$L_{cate}$ for semantic classification]{
\includegraphics[height=0.3322\textwidth]{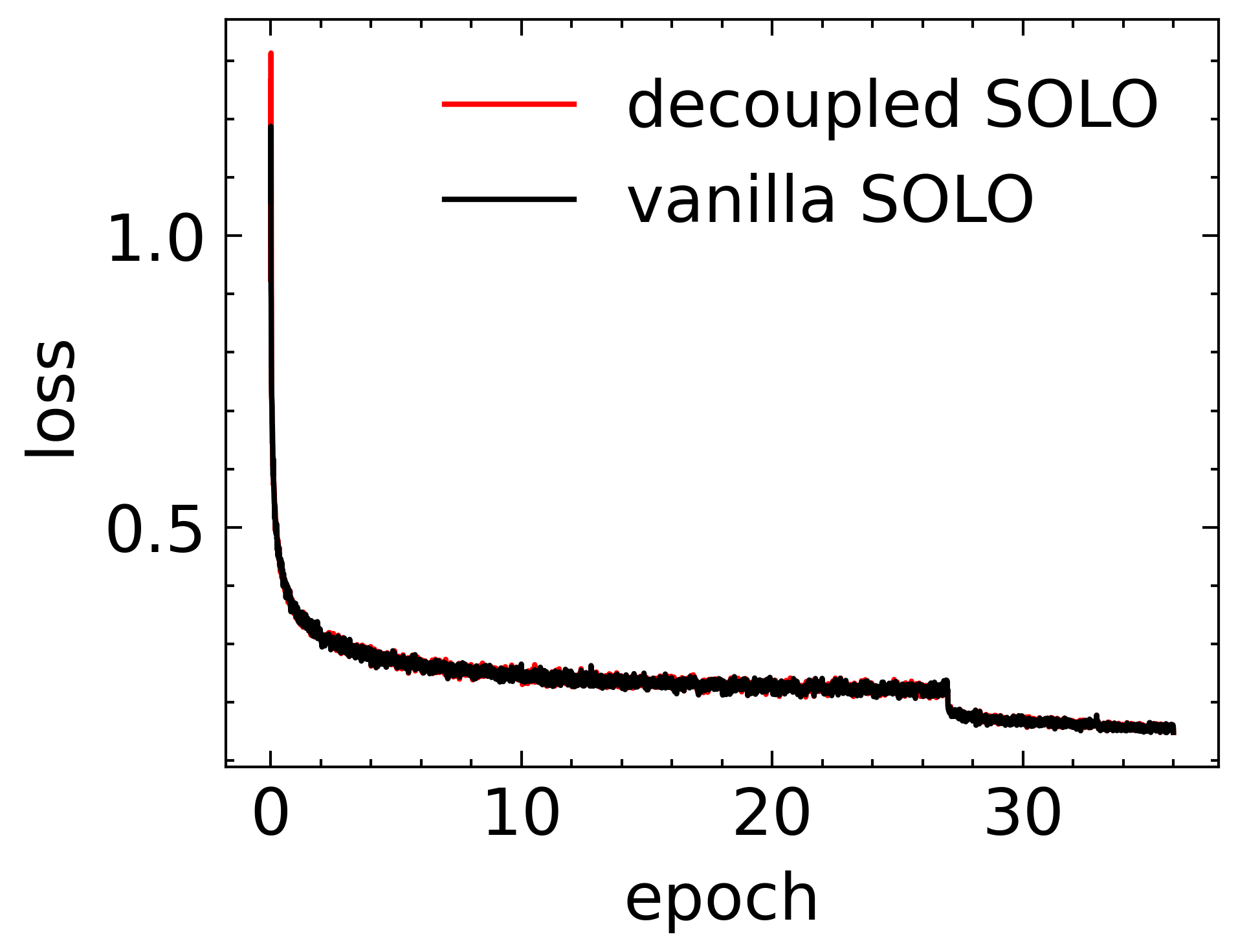}
\label{fig:loss_curve_cate}
}
\subfigure[$L_{mask}$ for mask prediction]{
\includegraphics[height=0.3321\textwidth]{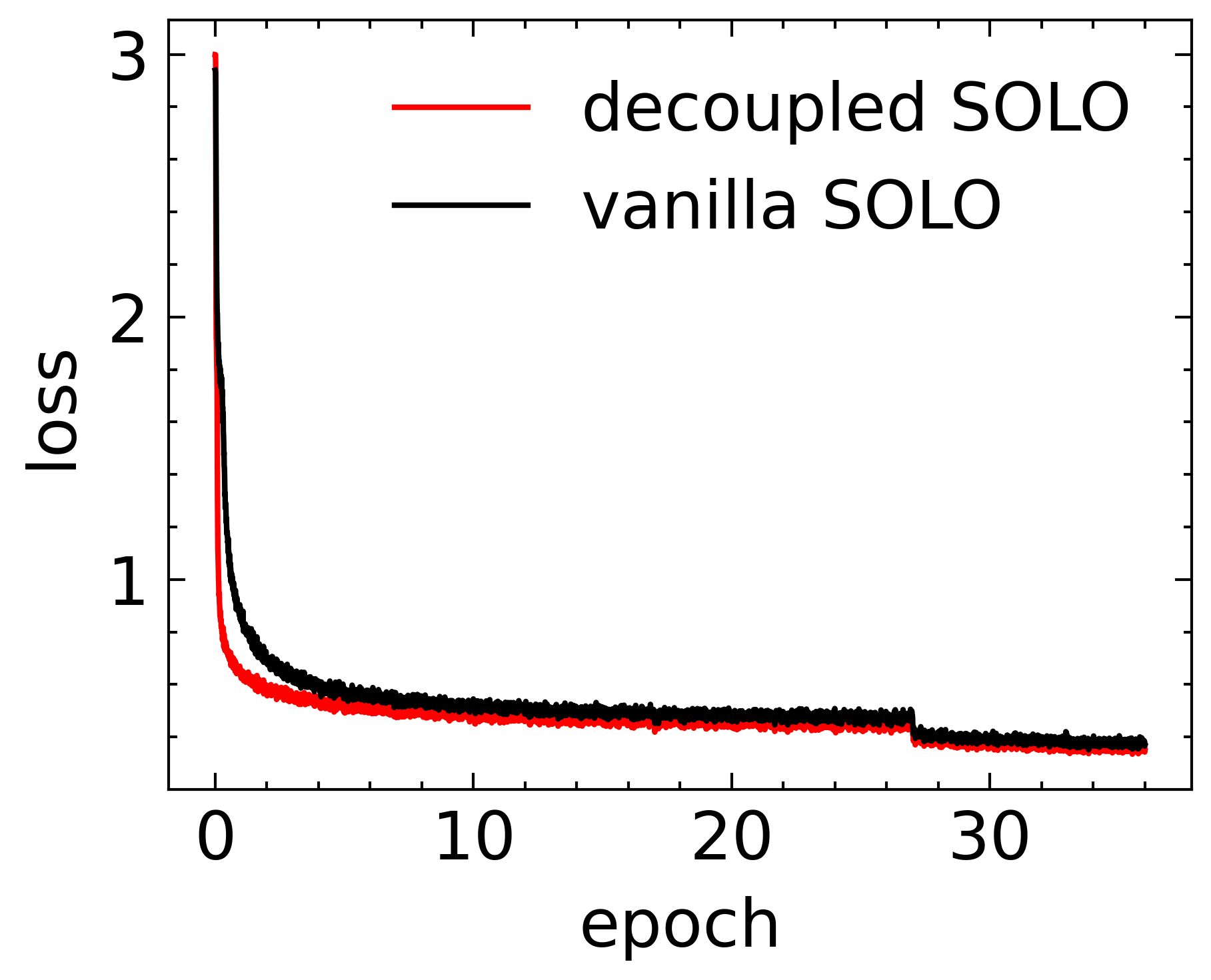}
\label{fig:loss_curve_mask}
}
\caption{Loss curves of the vanilla SOLO and decoupled SOLO. The models are
trained on MS COCO \texttt{train2017} with
36 epochs
and tested on \texttt{val2017}.}
\label{fig:loss_curves}
\end{figure*}

\end{document}